\documentclass[10pt,journal,compsoc]{IEEEtran}
\ifCLASSOPTIONcompsoc
  \usepackage[nocompress]{cite}
\else
  \usepackage{cite}
\fi
\ifCLASSINFOpdf
   \usepackage[pdftex]{graphicx}
\else
\fi
\usepackage{amssymb}

\usepackage{tabularx}
\usepackage{longtable}
\usepackage{makecell}
\usepackage{multirow}
\usepackage{multicol}
\usepackage{arydshln}
\usepackage{booktabs}
\usepackage{xcolor}
\usepackage{multirow}
\usepackage{arydshln}
\usepackage[shortlabels]{enumitem}
\usepackage{romannum}
\usepackage{subfig}

\usepackage{multirow}
\usepackage{amsmath}

\usepackage{listings}
\usepackage{xcolor}

\definecolor{myGreen}{RGB}{0, 70, 0}
\definecolor{orange}{RGB}{172,100,6}
\definecolor{mag}{RGB}{210,0,210}
\definecolor{red}{RGB}{225,0,0}
\definecolor{blue}{RGB}{0,0,225}

\newcolumntype{?}{!{\vrule width 1pt}}

\usepackage{url}
\usepackage{siunitx}

\usepackage{threeparttable}  

\usepackage{dblfloatfix}

\usepackage{comment}

\begin{document}
%
\title{Using Affect as a Communication Modality to Improve Human-Robot Communication in Robot-Assisted Search and Rescue Scenarios}
%
%
%
%

\author{Sami Alperen~Akgun,
        Moojan~Ghafurian,
        Mark~Crowley,
        and~Kerstin~Dautenhahn,~\IEEEmembership{Fellow,~IEEE}
\IEEEcompsocitemizethanks{\IEEEcompsocthanksitem S. A. Akgun is with the Department of Systems Design Engineering, University of Waterloo, Waterloo, ON, N2L 3G1, Canada.\protect\\
E-mail: saakgun@uwaterloo.ca

\IEEEcompsocthanksitem M. Ghafurian is with the Department of Electrical and Computer Engineering, University of Waterloo, Waterloo, ON, N2L 3G1, Canada.\protect\\
E-mail: moojan@uwaterloo.ca%

\IEEEcompsocthanksitem M. Crowley is with the department of Electrical and Computer Engineering, University of Waterloo, Waterloo, ON, N2L 3G1, Canada.\protect\\
E-mail: mcrowley@uwaterloo.ca

\IEEEcompsocthanksitem K. Dautenhahn is with the Departments of Electrical and Computer Engineering and Systems Design Engineering, University of Waterloo, Waterloo, ON, N2L 3G1, Canada.\protect\\
E-mail: kerstin.dautenhahn@uwaterloo.ca}

}

%
%

\markboth{}%
{Shell \MakeLowercase{\textit{et al.}}: Bare Demo of IEEEtran.cls for Computer Society Journals}
%



\IEEEtitleabstractindextext{%
\begin{abstract}

Emotions can provide a natural communication modality to complement the existing multi-modal capabilities of social robots, such as text and speech,  in many domains. We conducted three online studies with 112, 223 and 151 participants to investigate the benefits of using emotions as a communication modality for Search And Rescue (SAR) robots. In the first experiment, we investigated the feasibility of conveying information related to SAR situations through robots’ emotions, resulting in mappings from SAR situations to emotions. The second study used Affect Control Theory as an alternative  method for deriving such mappings. This method is more flexible, e.g. allows for such mappings to be adjusted for different emotion sets and different robots. 
In the third experiment, we created affective expressions for an appearance-constrained outdoor field research robot using LEDs as an expressive channel. Using these affective expressions in  a variety of simulated SAR situations, we evaluated the effect of these expressions on participants' (adopting the role of rescue workers) situational awareness. 
Our results and proposed methodologies provide (a) insights on how emotions could help conveying messages in the context of SAR, and (b) evidence on the effectiveness of adding emotions as a communication modality in a (simulated) SAR communication context.

\end{abstract}

}

\maketitle

\IEEEdisplaynontitleabstractindextext

%
\IEEEpeerreviewmaketitle

\IEEEraisesectionheading{\section{Introduction}\label{sec:introduction}}

%
%
%
%
\IEEEPARstart{E}mergency situations that require Search And Rescue (SAR) operations have been increasing on a yearly basis~\cite{feng2021review}. These situations may happen due to natural or man-made~\cite{shiri2020online} causes and require an immediate response, as time is a key element for the success of SAR operations~\cite{adams2007search}. Therefore, improving communication efficiency in SAR teams can be beneficial for the success of time-critical rescue operations.

The member composition of SAR teams has been changing over time. First, rescue dogs were included to help SAR teams by taking advantage of dogs' strong sense of smell, which can help find victims faster~\cite{slensky2004deployment}. More recently, rescue robots have become a part of SAR teams. Various rescue robots have been successfully employed in real SAR operations depending on the SAR type, such as snake robots~\cite{miller200213,hutson2017searching}, shape-shifting robots~\cite{ye2006design}, ground robots~\cite{Bethel2011,murphy2004trial}, drones~\cite{pratt2006requirements,michael2014collaborative}, or underwater vehicles~\cite{murphy2014disaster,matos2016multiple}. There are many reasons behind the widespread use of rescue robots in real-life scenarios, such as (a) SAR areas being unreachable or not safe for human rescuers due to various hazardous conditions such as extreme heat~\cite{casper2003human}, the toxicity of the environment~\cite{Gwaltney2003}, or confined spaces ~\cite{Linder2010}; (b) deploying robots to target SAR areas might be more time-efficient than deploying human rescue workers (thus increasing the operation's speed of progress); and (c) the limited number of human rescue workers since training human rescue workers requires a lot of time and effort~\cite{federal2013technical}.

Although rescue robots have been used in SAR operations since early 2000s~\cite{casper2003human}, they still need external help to operate appropriately. To the best of our knowledge, to date, there are no fully autonomous rescue robots or robot teams that can operate in unstructured and cluttered real-life SAR operations~\cite{delmerico2019current}. However, rescue robots can still act as teammates and improve human rescue workers' efficiency. To that end, a high level of collaboration between human-robot teammates should be achieved, which requires implementing clear and natural (i.e. intuitive)  communication channels between the human and robot teammates. However, human-robot interaction has been identified as a bottleneck in robot-assisted SAR operations~\cite{casper2003human,delmerico2019current}. In many situations, the intentions behind robot teammates' actions are not clear to the field workers, i.e., they do not know what the robot is doing or why it is behaving in a specific way. This lack of transparency in robot teammates' behavior has been identified as the main reason for inefficiency in SAR teams~\cite{kruijff2014experience}. Therefore, using affective communication between human field workers and rescue robots by taking advantage of multi-modal communication and developing alternative modalities for robot to human communication might help overcome this bottleneck in robot-assisted SAR operations.

Most of the rescue robots used today are already equipped with different communication modalities such as voice, text, photos, and videos~\cite{jones2020}. Nonetheless, these modalities may not always be effective in providing efficient communication in human-robot SAR teams. Generally, the specific situation that initiates search and rescue efforts affects the selection of suitable communication modalities. Other factors such as network traffic and the number of people using a specified radio frequency can also cause delays and/or miscommunications between SAR team members (e.g., see~\cite{SARExample}). Note, voice is often not effective in   SAR operations because rescue scenes tend to be noisy~\cite{Liu2006,Bethel2011}. Modalities other than voice can work in noisy environments, but they put the extra mental workload on field workers, or they do not work well depending on the search scene due to technical problems like delays (e.g., in understanding a message passed to rescue workers) and interferences (e.g., when a command does not contain the most recent information and inference needs to be made to predict the status)~\cite{jones2020} (see~\cite{jones2020} for more details on these situations and challenges). Hence, a combination of different communication modalities can help create a more robust communication in human-robot SAR teams, to ensure that if one of the modalities stops working accurately, the others can be used as alternatives. In other words, using multiple channels for conveying the same message can ensure effective operation.  In this article, we propose using affective expressions including emotions and moods (referred to as ``emotions" in the rest of the article) in a communicative way to complement existing communication modalities in human-robot SAR teams.

Many different theories exist that define emotions, such as Ekman's Psychoevolutionary Theory of Emotions~\cite{ekman1999basic}, James-Lange Theory~\cite{james1922emotions} or Cannon-Bard Thalamic Theory of Emotions~\cite{cannon1927james,Saraiva2019}. For example, based on Ekman's definition~\cite{ekman1999basic}, emotions are caused by a specific event. Ekman argues that basic emotions (sadness, happiness, fear, surprise, disgust, anger) are innate, present from birth, and universally recognized. Darwin also agreed on the universality of emotions and claimed that even people in isolated areas have similar emotional expressions~\cite{darwin1872expression}. Therefore, people are believed to be skilled at perceiving basic emotions without any training, and this process is believed to be intuitive, so it does not require significant mental workload~\cite{van2018}. This makes using emotions an excellent modality to complement the existing multi-modal communication methods used in SAR robots. Employing this modality could contribute to overcoming the present problems in SAR robots related to interaction among teammates (humans and robots). It could offer a way to reduce the cognitive load of human teammates to understand robot teammates' behavior during SAR  operations~\cite{Kolling2016}. Providing a way for robots to express emotions will also give SAR robots an ability to interact socially with humans, which could help rescue teams to operate in a more natural and efficient way~\cite{Bethel2011}. This social ability of robots has also the potential to help victims in SAR situations who encounter robots to feel calmer until the medical treatment team arrives, e.g., by preventing a shock~\cite{marx2013rosen,Bethel2011}, and is considered to be necessary for building affective robots that can communicate with humans more naturally~\cite{sefidgar2016}.

Despite all the work on implementing affective expressions for social robots, to the best of our knowledge, only one study attempted to use affective expressions on rescue robots~\cite{bethel2010non}. Bethel and Murphy suggested design guidelines to use body movements, postures, orientation, color, and sound to implement non-facial and non-verbal affective expressions on SAR robots, namely iRobot Packbot Scout and Inuktun Extreme-VGTV. They simulated a disaster site to conduct a user study to test the effectiveness of those suggested guidelines~\cite{bethel2010non}. While the guidelines were used to create a social robot (which was compared with a robot that did not have these capabilities)~\cite{bethel2010non}, a set of emotions which changed based on different SAR scenarios was not defined for the robot, which is what is investigated in our work. Furthermore, unlike Bethel and Murphy's work, we propose to use affective expressions as a \emph{complementary} communication modality to increase the efficiency of multi-modal human-robot communication in SAR teams. We believe such an approach can provide further insight into SAR robotics, emphasizing the need for interdisciplinary approaches.

Prior to implementing emotions for SAR robots, it is important to first study if it is feasible to convey SAR messages through emotions. Understanding whether there is a consensus in perception and expression of such emotions is necessary to verify whether communication through emotions would be possible and effective. Otherwise, it will not be clear what emotion a robot should show in a specific  situation, and this  might add the risk of miscommunication. In addition, it is important to study which emotion should a robot show in a specific SAR situation, to be able to add emotions as an additional communication modality.   

As our goal is to improve communication between robot-assisted SAR team members, this paper presents three online experiments that aim to understand (a) if affective expressions can be used for communicating SAR situations, (b) how a mapping between affective expressions and SAR situations can be obtained in a way that can be generalized for different robots with different affective expression abilities, and (c) if affect added as a complementary modality can in fact improve understanding of situations when other modalities fail. The primary motivation of these studies is to understand how emotions can be used as a complementary communication modality for robot to human communication in SAR teams, alongside other existing multi-modal methods.

The rest of the paper is organized as follows. Section \Romannum{2} presents our research questions. Related work is discussed in Section \Romannum{3}. Experiments 1, 2 and 3 with their corresponding results are explained in Sections \Romannum{4}, \Romannum{5}, and \Romannum{6}, respectively. Section \Romannum{7} presents a general discussion of the results, followed by a conclusion in Section \Romannum{8} and discussion of limitations \& future work in Section \Romannum{9}.

\section{Research Questions and Hypotheses}

In this article, we address the following research questions.

\begin{enumerate}
    \item [\textbf{RQ1}] Is there a consensus on what emotions should be used by Urban Search and Rescue (USAR) robots when they try to  convey information to human team members  about situations commonly occurring during USAR operations?
    \item [\textbf{RQ2}] Is the mapping between emotions and USAR situations robust and not dependent on the wording of the sentences?
    \item [\textbf{RQ3}] How can a mapping between SAR related sentences and emotions be obtained, and is there a way to generalize such mapping without limiting it to a specific set of emotions?
    
    \item [\textbf{RQ4}] Can affective expressions complement and improve multi-modal communication in human-robot SAR teams?
    \begin{enumerate}
        \item [\textbf{H1}] Affective expressions will increase participants' situational awareness (i.e. their perception of what is happening in the disaster area) when other communication modalities like text fail.
        
    \end{enumerate}
 
\end{enumerate}

\section{Related Work}

In this section, we will first introduce SAR and then describe then discuss the state of the art in the following areas that are relevant to our work: robots in SAR situations, research on Human-Robot Interaction (HRI) for existing SAR robots, using affective expressions in HRI, as well as some relevant work on sentiment analysis.

\subsection{Search and Rescue}
SAR is the general term for searching for people who are lost, trapped, and (might be) in danger. It is a broad term and has many sub-fields, usually depending on the search area, such as Mountain Rescue~\cite{karaca2018potential}, Cave Rescue~\cite{jackovics2016standard}, Urban Search and Rescue (USAR)~\cite{baker2004improved} and Wilderness Search and Rescue (WSAR)~\cite{goodrich2008supporting}. Regardless of the type of SAR, time is always a critical factor~\cite{adams2007search}. Thus, fast and efficient communication among SAR team members can be a deciding factor in whether or not the SAR team will succeed in saving people's lives.

\subsection{Robots in SAR}

The idea of using robots to assist SAR operations has been around since the early 2000s~\cite{1337826}. Initial research on SAR robots focused on the control of the robots, i.e., designing robust controllers to allow users to operate rescue robots easily~\cite{liu2013robotic,matsuno2014utilization}.

After successfully utilizing SAR robots, researchers shifted their focus to designing methods to reduce human teleoperators' workload. Low-level autonomous robot behaviors in SAR operations (e.g., the ability to climb up/down stairs autonomously without explicit human input) were designed~\cite{mourikis2007autonomous}. Semi-autonomous control methods were tested with adjustable autonomy levels in different scenarios, such as involving single robot-single operator~\cite{finzi2005human} or single operator-multiple robot teams~\cite{wegner2006agent}.

Machine learning (ML) techniques have been employed for robot-assisted SAR applications as well, e.g., to improve the efficiency of proposed controllers for SAR robots~\cite{doroodgar2010hierarchical}. More recently, researchers started to take advantage of ML methods to process sensory data that allowed SAR robots to better understand the rescue environments~\cite{niroui2019deep,lygouras2019unsupervised}.

To overcome the black-box nature of the majority of the existing ML methods, researchers advocated eXplainable Artificial Intelligence (XAI)~\cite{anjomshoae2019}. It has been argued that the explainability of  robots is needed to foster natural interactions~\cite{hellstrom2018}. Otherwise, human users might (a) not trust the robot when it takes a correct action but does not justify it, thinking that the robot's action might be due to an error~\cite{rossi2017timing}, or (b) assume that there is a logic behind every observed behavior of a robot while there may not be a clear logic and an action may rather be a result of an internal error in the robot's decision-making system~\cite{mirnig2017err}, underlying reasons for which could range e.g. from sensor errors, faulty actuators, software bugs, to incomplete or contradictory knowledge.

\subsection{Human-Robot Interaction in SAR}

Most of the research in the HRI field related to robot-assisted SAR have focused on improving teleoperation of SAR robots rather than on the interaction itself~\cite{kleiner2007mapping}. In addition, some studies investigated swarm robots for SAR applications, but their focus was still on how to reduce human teammates' cognitive load (e.g,~\cite{Kolling2016}). To the best of our knowledge, only a few studies have focused on interactions between human and robot teammates in SAR. Researchers in~\cite{chaffey2019developing} developed a virtual reality simulation for verbal communication in human multi-robot SAR teams, and they recorded data to create a better swarm emergency response where robots can clearly communicate with humans in the disaster area.

In another study, researchers analyzed the trade-off between the number of human operators and the number of rescue robots in a team for the Robocup rescue competition, taking operators' decision time and mental workload as optimization parameters~\cite{sato2004cooperative}. Further, a specific simulation environment for USAR (USARSim) was proposed and employed to reduce human operators' mental workload and stress levels~\cite{lewis2010teams}. A few studies have also focused on interactions between human and robot teammates in SAR. For example, in~\cite{kleiner2007mapping}, RFID tags in the SAR environment were used to exchange information between teammates to increase the mapping quality for gaining better team performance. 
Also, to simulate verbal communication in human multi-robot SAR teams, a virtual reality simulation was proposed in~\cite{chaffey2019developing} to create a better swarm emergency response where robots can clearly communicate with humans in the disaster area~\cite{chaffey2019developing}. Hada and Takizawa~\cite{hada2011} also showed promising outcome for remotely controlling rescue robots from a long distance (700m) using ad-hoc radio signals. Although it was not implemented, usage of gestures to communicate with search and rescue UAVs was proposed in~\cite{mayer2019}.

While focusing on these different aspects of communication, the work on the social side of HRI in robot-assisted SAR is quite limited. Fincannon et al. found out that rescue workers expect SAR robots to have social capabilities~\cite{fincannon2004evidence}. Furthermore, Murphy et al. surveyed 28 medical doctors and therapists who operated rescue robots to interact with victims~\cite{murphy2004robot}. They argued that it is important for rescue robots to have social capabilities to relieve victims until physical assistance arrives. They also stated that having social intelligence may contribute to building less ``creepy'' rescue robots~\cite{murphy2004robot}.

\subsection{Affective Expressions in HRI}

Although integrating emotions into SAR robots has not seen much attention, emotions, in general, have been one of the popular topics in HRI. Many HRI researchers have focused on how to use the embodiment of robots to express human-like emotions. Creating affective expressions in HRI has been investigated in multiple studies, e.g., using humanoid robots with a human-like embodiment (i.e.\ a head, face, arms, hands)~\cite{breazeal2003emotion,churamani2020icub}, or creating human-like expressions on non-human-like robots (e.g., using Action Units)~\cite{saldien2010expressing,sosnowski2006design}. Many studies  focused particularly on the utilization of facial expressions to implement expressive emotions for robots such as Kismet~\cite{breazeal2003emotion}, iCub~\cite{churamani2020icub}, and Probo~\cite{saldien2010expressing}. 
Other research uses animal-like (zoomorphic) robots. A a recent study described the design of affective expressions for the animal-like Miro robot,  based on a diverse literature on animals' and humans' non-verbal expression of emotions through head, face, and body movements. The obtained mapping, between affective expressions and their recognition  by human participants were found to be robust~\cite{ghafurianMiro}, even under visibility constraints~\cite{ghafurian2021recognition}, i.e., visibility situations similar to those that rescue workers also experience in SAR.

However, other, more machine-like and thus `appearance-constrained' robots pose different challenges for creating affective expressions that are legible to human interaction partners~\cite{Bethel2011}. Even for these types of robots, one of the few existing approaches suggested in the literature has been inspired by biological and ethological rules~\cite{korcsok2018biologically}. Related to SAR scenarios, in one study, researchers designed affective flight trajectories for drones to create expressive emotions for human users, taking inspiration from a performing arts method called the Laban Effort System~\cite{sharma2013communicating}. In addition to employing motions to implement affective expressions, color~\cite{rea2012roomba,kayukawa2017influence,collins2015saying}, sound~\cite{song2017expressing}, or touch~\cite{sefidgar2016,andreasson2018affective} (either individually or in combinations) were used to create more affective robots.

\subsection{Sentiment Analysis}

Many previous work on sentiment analysis have focused on mapping sentences with emotions, moods, or sentiments~\cite{Thelwall2012SentimentSD}. The classification of emotions in this process can be binary as in~\cite{turney2002thumbs}, where researchers categorized sentences as recommended (thumbs up) or not recommended (thumbs down) using unsupervised learning. Other work in this area goes beyond the mapping between sentences and emotions but tries to find the reason behind the predicted emotion (i.e., emotion stimuli). For example, in~\cite{ghazi2015detecting}, researchers trained a model to detect the best-associated emotion and its stimuli for given sentences.

There has also been some work on the intersection of sentiment analysis and HRI. Russell et al. (2015) took advantage of speech-to-text technologies to apply sentiment analysis on the conversation between humanoid robot MU-L8 and people interacting with it to improve the human-robot conversation~\cite{russell2015real}. Mishra et al. (2019) applied sentiment analysis methods to the feedback of customers interacting with the humanoid social robot called Nadine, to gain more insight on customers' expectations and how to use robots in real-world workplaces~\cite{mishra2019can}. Despite all the success obtained so far, the most significant limitation is that results highly depend on the context~\cite{nazir2020issues}. In other words, obtained mappings between text and emotions might differ drastically if the context of sentences changes (for example, a simple sentence like ``I see someone" expressed by a rescue robot would be perceived differently in a search for survivors scenario, as compared with a domestic security robot operating in someone's home at night.). Hence, mappings between sentences and emotions in the context of SAR may also be different from those that are currently suggested for the other contexts.

\subsection{Affect Control Theory}
Emotion prediction and modeling have been studied extensively by different research communities so far. Theories were introduced to classify emotions along several dimensions. Two well-known examples are the PAD emotional state model~\cite{mehrabian1995framework} and Affect Control Theory (ACT)~\cite{heise2007expressive}. These models use three dimensions: Pleasure, Arousal, Dominance (PAD) or Evaluation, Potency, Activity (EPA) dimensions, respectively, to describe the emotional meanings of words. Such dimensional emotion models usually have mappings that consist of ratings for different words (e.g., gathered through extensive surveys for EPA). Our study uses this method to decide on the mappings between situations in SAR and emotional expressions of a robot.

\section{Experiment 1}
\label{study1}

In this experiment, addressing \textbf{RQ1} and \textbf{RQ2}, we asked if it would be feasible to use emotions in SAR robots. In other words, we asked if there would be consensus in the mapping of SAR-related sentences to affective expressions~\cite{Akgun2020}. We also investigated whether such a mapping is robust to the wording of the sentences.

An online questionnaire was used where the participants were asked to select one or multiple affective expressions from a set of 11 expressions (including emotions and moods)\footnote{in this paper, all of these affective expressions are referred to as ``emotions''} that they believed could express a situation during USAR. The choices for the emotions were bored, sad, surprise, calm, disgust, angry, tired, annoyed, fear, happy, and excited. The situations were selected in a way that they represented ten common situations during USAR missions with two different wording styles (experimental conditions): \textit{social and intelligent conversational agent style} and \textit{system status report style} (see the first column in Table~\ref{table:final_results_exp1}). This was to ensure that the obtained mappings are not dependent on the wording of sentences/situations. We used this emotion set as it includes both basic and complex emotions. Moreover, this set was previously designed, implemented, and evaluated on a zoomorphic social robot~\cite{ghafurianMiro}, which expressed emotions through body movements, including head and face.

A questionnaire was also used to gather participants' (a) demographics information, (b) SAR experience, (c) perception of robots, and (d) self-reported ability to understand and show emotions (See~\cite{Akgun2020} for more details). We also had attention and consistency checks when showing situations (e.g., asking participants to select ``happy"), as well as in the questionnaire.

\subsection{Procedure}
First, participants reviewed and accepted the consent form and read the instructions for the study. Then, they read an example of an USAR scenario and were shown five different images of USAR robots in various shapes (machine-like, animal-like, or human-like). This was to help participants envisage the provided USAR scenario while not being biased by a specific robot's appearance. All images of the USAR robots were represented as black and white line drawings (see~\cite{Akgun2020} for an example of these images).

\begin{figure}[!t]
    \centering
    \captionsetup{font=scriptsize}
    \includegraphics[width=0.8\columnwidth]{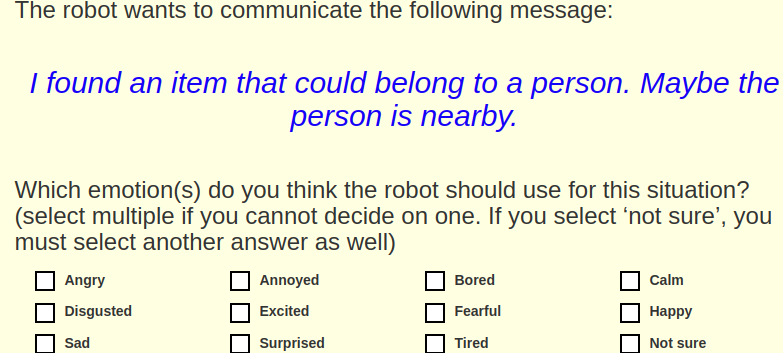}
    \caption{Interface Used in Experiment~1}
    \label{fig:interface_exp1}
\end{figure}

After reading the example USAR scenario and getting familiar with the concept, participants saw the USAR related statements in a random order, and they were asked to select one or multiple emotions that they thought would be appropriate for a robot to show in that situation (see Figure~\ref{fig:interface_exp1}). After the completion of mapping all ten sentences to emotions, participants answered the above mentioned questionnaire.

\begin{table*}[!t]
\begin{center}
  \scriptsize
  \captionsetup{font=scriptsize}
 \caption{Findings of the first experiment. Only emotions that were selected significantly more than random are included ($^{***}$: $p<.001$, $^{**}$: $p<.01$, and $^{*}$: $p<.05$). Two suggested mappings are shown in green and orange respectively. Cond. A: Condition Social and Intelligent Conversational Agent Style, Cond. B: Condition System Status Report Style } 
\begin{tabular}{| p{10cm} |c|l|l|l|}
 \hline
 \multirow{3}{*}{\bf Sentence} & \multirow{3}{*}{\bf Cond.} & \multicolumn{3}{c|}{\bf Significant Expressions} \\
 \cline{3-5}
 & & & \multicolumn{2}{c|}{\bf Suggested Mappings}\\
 \cline{4-5}
 & &  & \bf $1^{st}$ Mapping & \bf $2^{nd}$ Mapping\\
 \hline
  I can again communicate with our team outside of the building. & A & excited$^{*}$ & \bf \textcolor{myGreen}{happy}$^{***}$ & \bf \textcolor{orange}{calm}$^{***}$\\
 \hdashline
 Communication with external team restored. & B & excited$^{*}$ & \bf \textcolor{myGreen}{happy}$^{***}$ & \bf \textcolor{orange}{calm}$^{**}$\\
 \hline
 I lost communication with our team outside of the building so we are on our own now. & A & NA & \bf \textcolor{myGreen}{fear}$^{***}$ & \bf \textcolor{orange}{annoyed}$^{*}$\\
 \hdashline
 Communication with external team lost. & B & sad$^{*}$ & \bf \textcolor{myGreen}{fear}$^{**}$ & \bf \textcolor{orange}{annoyed}$^{**}$\\
 \hline
 I am stuck and might need help to proceed. & A & fear$^{**}$ & \bf \textcolor{myGreen}{annoyed}$^{***}$ & NA\\
 \hdashline
 Stuck here. & B & NA & \bf \textcolor{myGreen}{annoyed}$^{***}$ & NA\\
 \hline
 I detected dangerous material here, let's proceed carefully. & A & NA & \bf \textcolor{myGreen}{fear}$^{***}$ & NA\\
 \hdashline
 Dangerous material detected here. & B & surprise$^{*}$ & \bf \textcolor{myGreen}{fear}$^{***}$ & NA\\
 \hline
 I believe we are behind schedule. I also noticed it is getting dark and there is not much time left. & A &  NA & \bf \textcolor{myGreen}{annoyed}$^{***}$ & \bf \textcolor{orange}{fear}$^{***}$ \\
 \hdashline
 Behind schedule. It is getting dark. & B & tired$^{**}$ & \bf \textcolor{myGreen}{annoyed}$^{***}$ & \bf \textcolor{orange}{fear}$^{**}$\\
 \hline
 I found an item that could belong to a person. Maybe the person is nearby. & A & happy** & \bf \textcolor{myGreen}{excited}$^{***}$ &  \bf \textcolor{orange}{calm}$^{**}$\\
 \hdashline
 An object that might belong to a person was found. & B & surprise$^{**}$ & \bf \textcolor{myGreen}{excited}$^{***}$ & \bf \textcolor{orange}{calm}$^{***}$\\
 \hline
 My battery is running low and I will stop working soon. & A & sad$^{**}$ & \bf \textcolor{myGreen}{tired}$^{***}$ & \bf \textcolor{orange}{fear}$^{**}$\\
 \hdashline
 Battery is running low. & B & NA & \bf \textcolor{myGreen}{tired}$^{***}$ & \bf \textcolor{orange}{fear}$^{*}$\\
 \hline
 I think I found a surviving person. & A & surprise$^{*}$ & \bf \textcolor{myGreen}{excited}$^{***}$, \bf \textcolor{myGreen}{happy}$^{***}$ &NA\\
 \hdashline
 Possible living person detected. & B & calm$^{*}$ & \bf \textcolor{myGreen}{excited}$^{***}$, \bf \textcolor{myGreen}{happy}$^{***}$ &NA\\
 \hline
 I detected that there might be a risk of further collapse so we should only proceed with caution. & A & NA & \bf \textcolor{myGreen}{fear}$^{***}$ & NA\\
 \hdashline
 Further risk of collapse detected. & B & NA & \bf \textcolor{myGreen}{fear}$^{***}$ & NA\\
 \hline
 I think I heard someone is calling for help, we might have found a survivor. & A & surprise* & \bf \textcolor{myGreen}{excited}$^{***}$ & \bf \textcolor{orange}{happy}$^{***}$\\
 \hdashline
 Possible call for help detected. & B & NA & \bf \textcolor{myGreen}{excited}$^{***}$ & \bf \textcolor{orange}{happy}$^{*}$\\
 \hline
\end{tabular}
 \label{table:final_results_exp1}
\end{center}
\end{table*}

\subsection{Participants}
112 participants from North America (Canada and the USA) were recruited on Amazon Mechanical Turk for this study. Inclusion criteria for recruitment were having an approval rate of at least 97\% based on at least 100 HITS on Mturk. Data from the participants who failed the attention or consistency check questions were removed. We had a total of 78 participants (48 male, 29 female, 1 other; ages 20-72, avg: 35.7) who passed all the checks. 40 of them were in group A (they saw the sentences shown in the social and intelligent conversational agent style). The remaining 38 were in group B (they saw the sentences shown in the system status report style). Participants received \$2 upon completion of the study and a pro-rated amount based on the number of questions participants answered when they did not complete the study. This study received ethics clearance from the University of Waterloo's Human Research Ethics Board. 

\subsection{Results}

\subsubsection{Wording Style Results}
To analyze whether the change in the wording of the sentences affected the obtained mappings between situations and emotions, we compared the selected emotions for each sentence pair (condition A versus B) using Pearson's correlation coefficient~\cite{benesty2009pearson}. All sentence pairs (e.g., ``I am stuck and might need help to proceed'' and ``stuck here'') were significantly correlated ($0.78 \leq r \leq 0.99$). This suggested that the obtained mappings were robust and does not seem to be affected by the wording of the sentences. 

\subsubsection{Mapping Results}
Table \ref{table:final_results_exp1} presents emotions that were selected significantly more than random or/and more than the other emotions in each situation. We also provide suggestions for emotions to be used in each of these situations. Emotions shown in green are our first suggested choice for the mapping, emotions shown in orange are the second suggested choice, as alternative mappings (e.g., in case a robot is not capable of showing a specific emotion). These suggestions are (a) based on the agreement between two conditions and (b) significance levels.

\subsubsection{Questionnaire Results}

Results indicated that the majority of the participants believed that rescue robots are necessary and useful. Most of the participants stated that they were good at understanding and showing emotions. On the other hand, they were not entirely familiar with USAR or/and rescue robots (see~\cite{Akgun2020} for more details).

\section{Experiment 2}

This experiment addresses \textbf{RQ3}. Since Experiment~1 suggested that mapping emotions to situations in SAR is feasible, we now asked if there is a method to obtain these mappings in a way that (a) the mappings would not solely depend on a set of emotions (e.g., the 11 emotions shown to the participants in the previous study), and (b) the mapping process would have a potential to be automated in the future. Therefore, in this experiment, we study whether it is possible to use the three dimensions associated with emotions in the PAD emotional state model~\cite{mehrabian1995framework} or ACT~\cite{heise2007expressive} (EPA). As these three dimensions are very similar in the two models, we decided to use the EPA dimensions (i.e., Evaluation, Potency, and Activity) of the ACT, as there exists datasets, mapping emotions and EPA dimensions, which were gathered through large surveys and have been updated over the years to account for possible changes over time, as well as including different countries, to account for potential cultural differences. 

While the Evaluation (E) dimension in ACT shows how ``good" an emotion, identity, action, etc., is, the Potency (P) dimension shows how ``powerful" something is, and the Activity dimension (A) shows how ``active" it is. For example, the EPA value for the emotion ``happy" is [3.44, 2.93, 0.92]\footnote{Note that EPA values are commonly rated in a range between $-4.3$ and $4.3$}, based on the U.S.A. 2015 Dataset~\cite{smith2016mean}, which is used in this experiment. This suggests that ``happy" is believed to be quite good, somehow powerful, and slightly active.

\begin{figure}[htbp]
    \centering
    \captionsetup{font=scriptsize}
    \includegraphics[width=0.8\columnwidth]{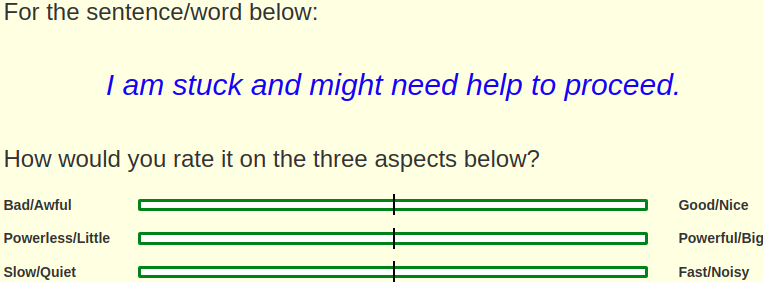}
    \caption{Interface Used in Experiment~2}
    \label{fig:interface_exp2}
\end{figure}

To study whether using the EPA dimension can lead to similar mappings,
instead of directly mapping sentences to a set of emotions (as in
Experiment~1), we asked participants to rate the sentences used in
Experiment~1 on the EPA dimensions (see
Figure~\ref{fig:interface_exp2}). Afterwards, we calculated the
emotion, from the set of 11 emotions used in Experiment~1, that was
the closest to the EPA rating for each sentence. 

\textbf{Selection of additional sentences:} Also, compared to Experiment~1, we included more sentences related specifically to different types of SAR, in addition to the sentences related to USAR used in the first experiment, to get insights on the potential validity of such mappings (i.e., to see if meaningful mappings can be obtained) for an extended set of situations (see sentences in row 11 to 16 in Table~\ref{table:mapping_exp2}). Since Experiment~1 suggested that the mappings were consistent and not affected by different wording styles, we only focused on sentences conveyed in the more expressive social and intelligent conversational agent style.

Table~\ref{table:mapping_exp2} shows the different sentences used in this study. Sentences (11) and (12) were included since there generally operates more than one field team in a search area, and the need for additional members might change dynamically depending on the given task~\cite{jones2020}. Sentences (13) and (14) were included because detecting the environment’s temperature is crucial for SAR sub-types involving extreme environments such as  deserts, in water, or with very cold climates. That is because the survival rate of victims decreases significantly both in cold water during maritime search and rescue~\cite{rafferty2013autonomous}, and in hot weather due to dehydration during WSAR~\cite{hung2007search}. Furthermore, sentence (15) was added since there is a chance to encounter an injured victim in SAR operations~\cite{jones2020,mu2020optimization,allouche2010multi,karaca2018potential}. Lastly, sentence (16) represents another scenario that is common during almost all SAR operations since it is usually impossible to directly reach some area of interest in the rescue field~\cite{jackovics2016standard,van2017wilderness,goodrich2008supporting,rafferty2013autonomous}. 

\textbf{Rating the sentences:} 
Participants were shown all the sentences used  in Experiment~1 and the additional sentences as in Table~\ref{table:mapping_exp2} in a random order, and they were asked to rate these sentences according to the EPA dimensions~\cite{heise2007expressive}. In other words, participants were asked to rate, on a continuous scale, how good, how powerful, and how active each sentence (and the corresponding situation it conveys) was (see Figure~\ref{fig:interface_exp2}).

As a consistency check, we asked participants to rate the words ``angry", ``good,” ``infant,” and ``boss" in addition to the sentences to compare these ratings with the original EPA values obtained from the U.S.A. 2015 Dataset~\cite{smith2016mean}. These words were selected as they cover a range of different values on each of the E, P, and A dimensions and could help ensure consistency between participants' ratings and the reference ratings in the dataset (used to find the closest emotion). In addition to these sentences and words, attention checks were included that instructed participants to select a specific answer (e.g., ``I found that for this sentence you have to select the leftmost option on all bars.").

\textbf{Obtaining the associated mappings:} Obtained EPA ratings were used to identify the corresponding emotion among the list of 11 affective expressions used in Experiment~1. The EPA ratings of the 11 emotions (from the most recent, U.S.A. 2015 Dataset~\cite{smith2016mean}) was compared with the obtained EPAs for each sentence. Euclidean distance~\cite{danielsson1980euclidean} was used to find the closest mapping. As an example, for the sentence ``I think we need additional team members'', we compared the distances between participants' EPA ratings for this sentence (e.g., [0.83,0.77,0.71]) and EPAs of all 11 emotions. We found that the closest distance (1.68) corresponded to the emotion ``surprised'' ([1.42,1.35,2.17]). 

\subsection{Procedure}

Participants first read the consent form and the instructions for the study. Afterward, they rated sentences (along with consistency and attention checks) on the EPA dimensions in a random order (see Figure~\ref{fig:interface_exp2}). Finally, they received an end code for the completion.

\subsection{Participants}
We recruited 223 participants (79 from Canada and 144 from the USA) on Mechanical Turk for this study. We started with the same recruitment criteria as in Experiment~1: having an approval rate of at least 97\% based on at least 100 HITS on Mturk, but later changed the criteria to an approval rate of 96\% based on at least 50 HITS on Mturk for participants who were from Canada~\footnote{None of the participants from Canada failed any of the attention checks, so we changed the criteria to be able to recruit more participants}. After filtering, based on the attention check questions, 133 participants remained (72 from Canada and 61 from USA). Participants were paid 0.3\$ for participation in this study. This study received ethics clearance from the University of Waterloo's Human Research Ethics Board. 

\begin{table*}[!t]
  \centering
  \captionsetup{font=scriptsize}
  \caption{Predicted emotions based on the distance between average EPA ratings of participants' EPA scores and EPA ratings of the 11 emotions from the U.S.A 2015 dataset~\cite{smith2016mean}. Corr shows the Pearson's correlation coefficient~\cite{benesty2009pearson} between EPA ratings of participants from Canada and USA. Dist. refers to the calculated Euclidean distance~\cite{danielsson1980euclidean} between EPA values of the predicted emotions and participants' ratings. The last column shows the mappings obtained in Experiment~1 for comparison.}
 \scriptsize
 \begin{tabular}{|p{0.25cm}|p{5.3cm}|l|l|l|l|l|l|p{0.9cm}|p{1.2cm}|p{2.0cm} |}
 \hline
  \multirow{2}{*}{\bf No} &
 \multirow{2}{*}{\bf Sentences} & \multicolumn{3}{c|}{\bf Average} & \multirow{2}{*}{\bf Corr} & 
 \multicolumn{2}{c|}{\textbf{ $1^{st}$ Prediction} } & %
    \multicolumn{2}{c|}{\textbf{ $2^{nd}$ Prediction} } & \multirow{2}{*}{\textbf{From Exp. 1} }\\
\cline{3-5}
\cline{7-10}
    & & \bf E & \bf P  & \bf A &  &\bf Dist. & \bf Emotion & \bf Dist. & \bf Emotion & \\
 \hline
   1 & I can again communicate with our team outside of the building & 2.55	& 1.73 & 0.93 &	0.98 & 1.39 & Excited	& 1.49 & Happy	& excited, \textbf{\textcolor{myGreen}{happy}}, \textbf{ \textcolor{orange}{calm}}\\
 \hline
   2 & I lost communication with our team outside of the building so we are on our own now. & -2.00 &	-1.35 &	-0.42 &	0.99 & 0.56 & Fearful & 1.23 & Annoyed & \textbf{\textcolor{myGreen}{fearful}}, \textbf{\textcolor{orange}{annoyed}}\\
 \hline
   3 & I am stuck and might need help to proceed  & -1.08 & -1.46 & -0.13 & 0.96 & 1.48 & Fearful & 1.49 & Annoyed	& fearful, \textbf{\textcolor{myGreen}{annoyed}} \\
 \hline
   4 & I detected dangerous material here, let's proceed carefully & -1.03 & 0.98 &	-0.56 &	0.88 & 1.76 &	Disgusted	& 2.17, \textcolor{mag}{2.43} & Annoyed, Fearful	& \textbf{\textcolor{myGreen}{fearful}}\\
 \hline
  5 & I believe we are behind schedule. I also noticed it is getting dark and there is not much time left & -1.72 &	-0.98 &	0.12 &	0.99 & 0.68 & Annoyed &	1.06 & Fearful &	\textbf{\textcolor{orange}{fearful}}, \textbf{\textcolor{myGreen}{annoyed}}\\
 \hline
   6 & I found an item that could belong to a person. Maybe the person is nearby & 2.16 &	1.23 & 	0.38 & 0.99 & 1.94 &	Surprised	&	2.16 & Excited	& happy, \textbf{\textcolor{myGreen}{excited}},  \textbf{\textcolor{orange}{calm}}\\
 \hline
   7 & My battery is running low and I will stop working soon & -1.80 & 	-1.73 &	-0.82 & 0.98 & 0.90 &		Fearful	&	1.35 & Sad	& sad, \textbf{\textcolor{myGreen}{tired}},  \textbf{\textcolor{orange}{fearful}}\\
 \hline
   8 & I think I found a surviving person & 3.00 &	2.63 &	1.70 &	0.96 & 0.77 & Excited & 0.94 & Happy	& surprised, \textbf{\textcolor{myGreen}{excited}}, \textbf{\textcolor{myGreen}{happy}} \\
 \hline
   9 & I detected that there might be a risk of further collapse so we should only proceed with caution & -0.99 &	0.37 &	-0.47 &	\textcolor{blue}{0.42} & 1.57 & Disgusted & 1.75, \textcolor{mag}{1.99} & Annoyed, Fearful	& \textbf{\textcolor{myGreen}{fearful}}\\
 \hline
  10 & I think I heard someone is calling for help, we might have found a survivor & 2.81 &	2.32 &	2.32& 0.98 & 0.20&	Excited& 1.65		&Happy	& surprised, \textbf{\textcolor{myGreen}{excited}}, \textbf{\textcolor{orange}{happy}}\\
 \hline
 11 & I think we need additional team members & 0.83 &	0.77 &	0.71 &	\textcolor{blue}{-0.29} & 1.68& Surprised & 2.79 &		Excited	 &NA\\
 \hline
 12 & I think we have more team members than we need. One of us should join the other team & 0.65 &	0.88	& 0.83 & \textcolor{blue}{0.22} &	1.61 & Surprised & 2.62 &		Angry&	NA\\
 \hline
 13 & I detected that the temperature of the environment is too cold for a person & -1.25 &	-0.30 &	-0.52 & 0.99 & 1.35 & 		Fearful & 1.36 & Annoyed &	NA\\
 \hline
 14 & I detected that the temperature of the environment is too hot for a person & -1.45 &	0.14 &	0.36 & 0.99 & 0.83 &	Disgusted & 0.97		& Annoyed	& NA\\
 \hline
 15 & I found a victim that requires medical attention & 0.06 &	1.42 &	1.88 &	0.97 & 1.39&	Surprised & 2.02	&	Angry &	NA\\
 \hline
 16 & I detected that this rescue route requires obstacle clearance & -0.25 &	0.49 &	0.54 &	0.94 & 1.98 &	Angry	& 2.04 &	Disgusted	& NA\\
 \hline
\end{tabular}
\label{table:mapping_exp2}
\vspace{-0.5cm}
\end{table*}

\subsection{Results}
In this section, we will first discuss how consistency checks were applied and will then present the results for the ratings and the obtained mapping between  situations and emotions.

\subsubsection{Consistency Checks}
Despite having a high approval rate criteria for recruitment on MTurk, 90 participants failed either or both of the attention check questions. Since attention check questions were related to selecting the right or left-most part of the bars, we included an error margin during the filtering. We accepted a range of answers that were not too far from the correct answer on the continuous scale (i.e., a 10\% error margin for both left and rightmost part of the continuous scale).

After removing participants who failed the checks, the mean EPA values of participants from the two countries we recruited from, i.e. USA and Canada, were compared to investigate whether/how cultural differences affected the ratings. As shown in Table~\ref{table:mapping_exp2}, we found high correlations between the ratings from the two countries, suggesting that the ratings were overall consistent between Canada and the USA. Therefore, the merged data was used for analyzing the results.

\begin{table}[!b]
  \centering
  \captionsetup{font=scriptsize}
  \caption{Comparison of mean EPA ratings scored by participants with EPA ratings obtained from the U.S.A. 2015 Dataset~\cite{smith2016mean}. Results of Pearson's correlation coefficient are shown in the last column.}
  \scriptsize
  \begin{tabular}{|l|l|l|l|l|l|l|l|}
  \hline
 \multirow{2}{*}{\bf Words} & \multicolumn{3}{c|}{\bf From Participants} & 
 \multicolumn{3}{c|}{\bf From Dictionary}& 
    \multirow{2}{*}{\textbf{Corr} }\\
    \cline{2-7}
    & \bf E & \bf P  & \bf A & \bf E & \bf P  & \bf A & \\
    \hline
    angry & -3.08 & 1.44 & 2.23 & -1.77 & 0.57 & 1.80 & 0.98\\
    \hline
    good & 3.69 & 2.32 & 0.45 & 3.40 & 2.37 & -0.24 & 0.99\\
    \hline
    infant & 2.62 & -2.45 & -0.43 & 2.26 & -2.35 & 1.23 & 0.91\\
    \hline
    boss & 0.77 & 3.07 & 1.61 & 0.91 & 2.79 & 1.07 & 0.96\\
    \hline
\end{tabular}
\label{table:scaling_exp2}
\vspace{-0.5cm}
\end{table}

\subsubsection{Scaling}

As we used a specific EPA dataset to find the closest emotion to each of the sentences, we first had to ensure that participants' ratings were consistent with those in the dataset. Therefore,
we first checked participants' ratings of the above mentioned words (i.e., angry, good, infant, and boss).  Averages of these EPA ratings were calculated and compared with EPA ratings obtained from the U.S.A. 2015 Dataset~\cite{smith2016mean} using Pearson's correlation coefficient~\cite{benesty2009pearson}. We found high correlations (see the last column of Table \ref{table:scaling_exp2}), which suggested that there would be no need to scale the obtained EPA ratings given by the participants in order to create the mappings.

\subsubsection{Mapping Results}

The results for EPA ratings, as well as the mapping outcomes, are shown in Table~\ref{table:mapping_exp2}. Each row in Table~\ref{table:mapping_exp2} contains the mean EPA values for a particular sentence and the two closest predicted emotions, calculated through the above-mentioned method (i.e., by comparing the Euclidean distances between mean EPA ratings and EPAs of the 11 emotions according to the U.S.A. 2015 Dataset~\cite{smith2016mean}). For each of the predicted emotions, the calculated distance (dist.) is stated. We also show the results from Experiment~1 in the last column for comparison. For example, for the sentence ``I can again communicate with our team outside of the building'', participants' average EPA ratings were: $E=2.55, P=1.73, A=0.93$. The correlation between the ratings in the USA and Canada was $0.98$, the closest emotion to average EPA ratings of participants was calculated to be ``Excited" with a distance of $1.39$, compared to EPA ratings in the dataset, and the second closest emotion was ``Happy" with a distance of $1.49$. These results were consistent with the mappings obtained from Experiment~1 (i.e., Excited, Happy, and Calm). Only for two sentences, the two closest emotions did not match with the ones obtained through the first experiment, but the third closest emotion matched. The third closest distance for these two sentences is shown in pink colour in Table~\ref{table:mapping_exp2}.

We also further analyzed the correlations between EPA ratings of participants from Canada and the USA. While the ratings were generally highly correlated for most situations, we observed that the correlation value was lower for three of the sentences ($-0.29$, $0.22$, and $0.42$). These sentences with lower correlation are shown in blue in Table~\ref{table:mapping_exp2}. Mean EPA ratings of these sentences for each country are presented in Table~\ref{tab:low_correlations}.

\begin{table}[!t]
    \centering
    \captionsetup{font=scriptsize}
    \scriptsize
    \caption{The sentences which have either low or no correlation between participants from Canada and U.S.A.}
    \begin{tabular}{|p{4.7cm}|l|l|l|l|}
    \hline
    \textbf{Sentence} & \textbf{Loc.} &  \bf (E) & \bf(P) & \bf(A)\\ 
    \hline
    \multirow{2}{4.7cm}{I detected there might be a risk of further collapse so we should only proceed...} & CA & -0.83 & 0.46 & -1.04\\
    \cline{2-5}
    & USA & -1.18 & 0.26 & 0.20\\ 
    \hline
    \multirow{2}{4.7cm}{I think we need additional team members} & CA & 0.58 & 0.66 & 0.54\\
    \cline{2-5}
    & USA & 1.12 & 0.89 & 0.92\\ 
    \hline
    \multirow{2}{4.7cm}{I think we have more team members than we need. One of us should join the other...} & CA & 0.48 & 0.61 & 0.74\\
    \cline{2-5}
    & USA & 0.85 & 1.21 & 0.94\\ 
    \hline
    \end{tabular}

    \label{tab:low_correlations}
\end{table}

\section{Experiment 3}
As Experiments~1 and~2 supported the feasibility of using affective expressions with SAR robots, in this third experiment we explored the effect of affective expressions on robot to human communication in the context of SAR teams to address research question \textbf{RQ4} and our hypothesis \textbf{H1}. In other words, we asked if affective expressions, used as an additional communication modality, can improve accuracy of communication in situations where other modalities may fail.

\subsection{The Husky Robot and Affective Expressions}
\begin{figure}[!t]
    \vspace{-0.5cm}
    \centering
    \captionsetup{font=scriptsize}
    \subfloat[Happy]{\includegraphics[width=0.49\columnwidth]{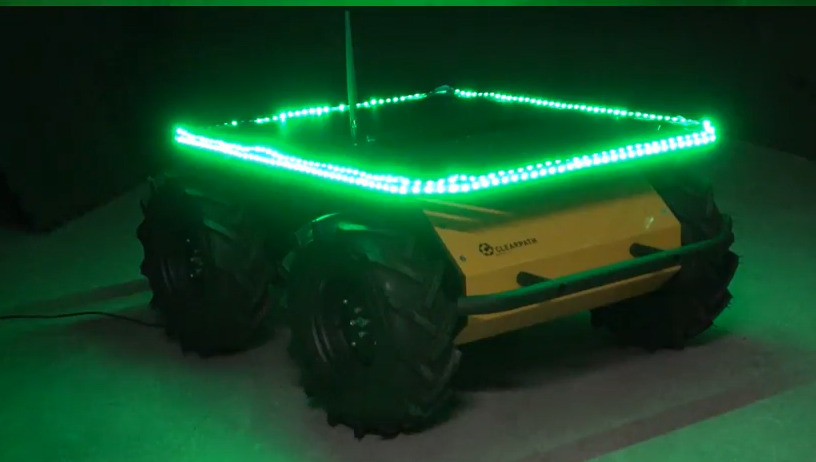}} \hfill
    \subfloat[Fearful]{\includegraphics[width=0.45\columnwidth]{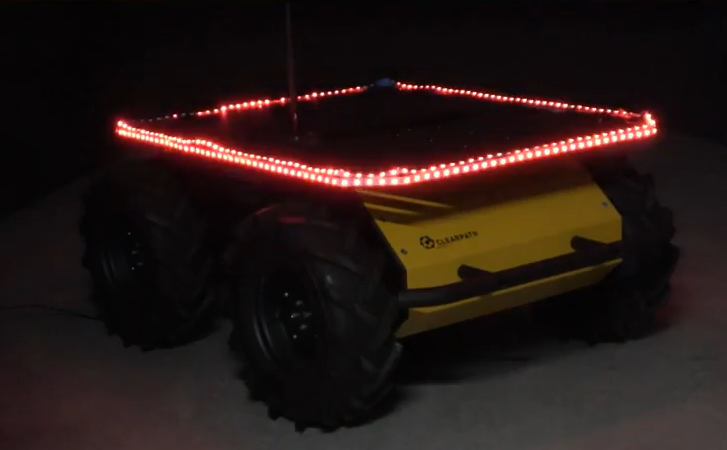}}\\
    \caption{Examples of designed light patterns}
    \label{fig:study4-lightsPilots}
\end{figure}

To be able to use affective expressions in scenarios with a robot that might realistically be used in SAR scenarios, we designed and implemented the expressions on Clearpath's Husky robot\footnote{https://clearpathrobotics.com/husky-unmanned-ground-vehicle-robot}, an appearance-constrained  robot. Affective cues were displayed using light signals (LED strips), based on EPA dimensions of ACT~\cite{heise2007expressive}. A series of informal pilot studies were conducted with recorded videos of Husky's affective expressions, and with lab members who were not involved in this research and who were not part of the participants that were subsequently recruited for the experiment. In each pilot, different parameters of these expressions (e.g., light intensity, frequency, patterns, etc.) were used. After analyzing the results of the pilot studies, we decided to continue with a full light pattern (i.e., turning on/off all the LEDs on a strip around the robot) since in this case the light patterns can more easily be observed from all viewing angles. In addition, we decided to use the \textit{color} of the lights for the Evaluation dimension (how good the emotion is), while using the \textit{brightness} of the lights to represent how powerful the emotion is (Potency), and the frequency of the light changes to represent the Activity dimension (see Figure~\ref{fig:study4-lightsPilots}). These design suggestions were also inspired by the work of Collins et al. (2015)~\cite{collins2015saying}.

\begin{figure}[!b]
    \centering
    \captionsetup{width=1.0\columnwidth, font=scriptsize}
    \subfloat[First Video]{\includegraphics[width=1.0\columnwidth]{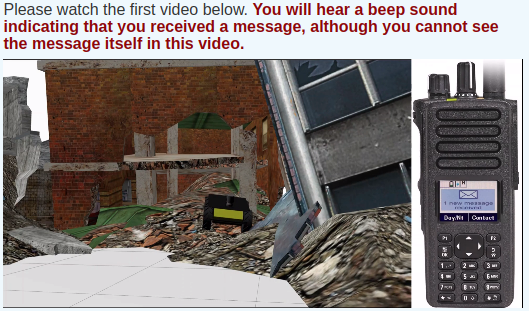}} \\
    \vspace{-0.3cm}
    \subfloat[Second Video (Emotion condition since Husky's lights are active)]{\includegraphics[width=1.0\columnwidth]{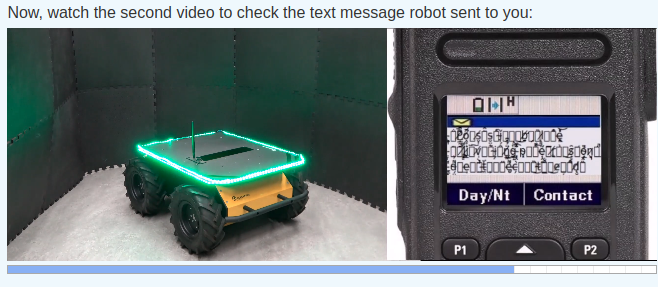}}\\
    \caption{The main task in Experiment 3 where participants were shown two videos for each scenario and then asked to guess what Husky wants to convey.}
    \label{fig:study4-mainTaskBothVideosandQuestion}
\end{figure}

After implementing affective light displays, noisy text messages were created to mimic a situation when the text communication modality fails during the SAR mission. Zalgo text with different chaos levels was used to distort the text messages and make them difficult to read~\cite{castano2013definingZalgo}. Here, again, a pilot study with different noise levels was conducted with lab members to decide on the noise level (one that would not be too easy to read/understand). The noisy text messages were presented to participants of the study as displayed on a radio transmitter device (Motorola XPR 7550e), which is widely used during real SAR missions~\cite{motorola7550}. Figure~\ref{fig:study4-mainTaskBothVideosandQuestion} shows an example of the noisy text message shown to participants.

The experiment had a between-participant design where participants were randomly divided into two conditions: \textit{Emotion} condition and \textit{No emotion} condition. While Husky expressed emotions using lights in the Emotion condition, it did not use emotions as a communication modality in the No Emotion condition (i.e. when no lights, and thus affective expressions, were displayed).

\subsection{Procedure}
The study consisted of three steps as below:

\textbf{Step 1:} Participants first completed an initial training step to learn the affective expressions of Husky. The training was similar to the training in~\cite{ghafurian2021recognition}. Participants first watched a training videos showing emotional displays along with a text indicating the corresponding emotion (this video could be played as many times). Then they were tested on their recall of these emotions, to see how well they learned the meaning of each of these emotional displays. Five emotions were used in this study: happy, excited, tired, annoyed, fear. This step was added to make sure that the participants in the emotion condition will know the meaning of emotional displays of Husky. This is also an expected step if emotions might in future be used in SAR robots as a communication modality, since, in practice, SAR workers get training regularly, including on how to use new tools. For such high-risk tasks such as SAR, one cannot rely on SAR team members being able to `intuitively' recognize the affective expression that they encounter for the first time, they would rather expect to be trained, to be able to operate and interact with the robot reliably. Thus,  adding this step to their regular training routine might actually be feasible in future applications.

\textbf{Step 2:} Each participant watched 20 videos for 10 different SAR scenarios (two videos for each scenario) and were asked to select what they thought the robot was communicating to them. Each scenario consisted of two videos (one providing the SAR context and one showing the message rescuers would receive as described below) which were shown to participants in a random order. After watching the two videos for each scenario, the participants were asked to select the message that they believed the robot wanted to convey to them. The list of messages to select from included all 10 different messages. It was a multiple choice option, so participants could select more than one message from the list. They could replay the videos as many times as they wished, in order to make a decision. The videos were automatically paused if they switched the interface tab or opened another application to measure their response time accurately (and to ensure participants  paid sufficient attention to the videos).

For each scenario, participants were shown 2 videos. The first video showed movements of Husky in a 3D simulated disaster environment and the second one showed the message that Husky intended to send to the participants, i.e., it showed the noisy message with/without emotional lights on the robot (depending on the condition) for each scenario. The participants were then asked to select the message that they thought Husky wanted to convey to them (see Figure~\ref{fig:study4-mainTaskBothVideosandQuestion}). The emotion to show in each situation was decided according to the results of Experiments~1 and~2.

\begin{figure}
    \centering
    \includegraphics[width=0.9\columnwidth]{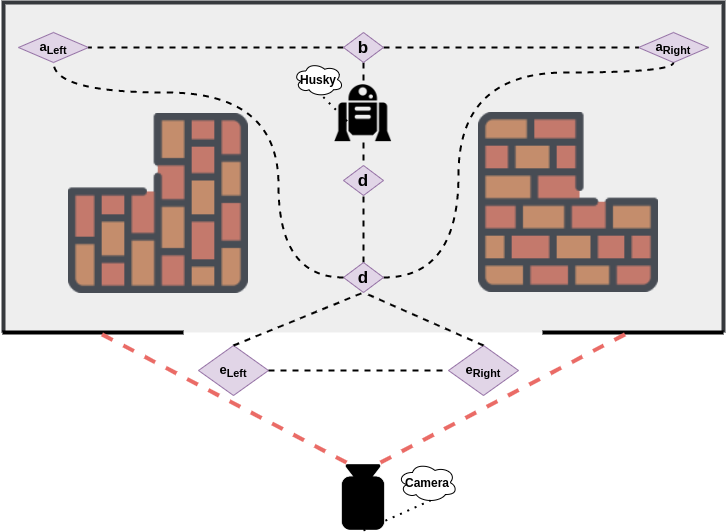} 
    \captionsetup{font=scriptsize}
    \caption{Experimental Setup used to mimic common scenarios happening during SAR missions in Experiment 3. Different locations were shown with pink rhombuses and labeled with letters. Possible routes between these points were shown with dashed lines. For each scenario, Husky starts its movement from one of these points and visits particular points using dashed routes.}
    \label{fig:study4-expSetup}
\end{figure}

\begin{table}[!b]
\captionsetup{font=scriptsize}
\caption{Different routes were followed by Husky for each scenario indicated by a number (corresponding scenario names are given in Table~\ref{table:mapping_exp2}). Please see Figure~\ref{fig:study4-expSetup} for experimental setup and locations of labelled points ($a_{Right}, a_{Left}, b, c, d, e_{Right}, e_{Left})$. Matched emotions for each scenario can be found in Table~\ref{table:final_results_exp1}.}
\label{tab:study4-expSetupScenarioMovements}
\footnotesize
\centering
\begin{tabular}{|c|c|}
\hline
\textbf{No} & \textbf{Path Followed}\\ \hline
1                                                  & $a_{Right}->b->c->d$                                       \\ \hline
2                                                 & $b->d->e_{Right}->e_{Left}->d$                                  \\ \hline
3                                                                     & $b->c->d->c->d->c->d$                          \\ \hline
4                                              & $b-a_{Left}->b->c->d$                                  \\ \hline
5      & $b->c->d$       \\ \hline
6                                & $b->a_{Right}->b->c->d$                         \\ \hline
7                                                      & $a_{Left}->b->c->d$ (decrease speed gradually)            \\ \hline
8                                                                        & $b->a_{Right}->d$ (use curvy path from a to d)     \\ \hline
9            & $b->a_{Left}->d$ (use curvy path from a to d)              \\ \hline
10                               & $a_{Right}->d$   (use curvy path from a to d)               \\ \hline
\end{tabular}
\end{table}

The experimental setup shown in Figure~\ref{fig:study4-expSetup} was used to simulate common scenarios happening during SAR operations and to give participants the context that search and rescue robots get in SAR operations. In this setup, Husky starts from a particular point depending on the scenario and follows a specific route. For example, for the scenario ``I think I found a surviving person'', Husky starts from point $b$ in Figure~\ref{fig:study4-expSetup}, then goes to point $a_{Right}$. During this movement, Husky slowly goes out of the view for the observing participant. Then, it appears again and moves toward point $d$ by taking the curvy path (shorter path comparing to going first point $b$ and then point $d$). 

At some point during these movements, participants were also notified that they got a text message from Husky, both visually (by seeing a text saying that they have received a message), and through sounds (a beeping sound similar to the receipt of a text message). The message was not shown to them at this point to make sure that they will focus on the movements of the robot and will consider the context. The time of receiving this text notification was controlled and differed in each scenario, to make it realistic. Different paths followed by Husky for each scenario are summarized in Table~\ref{tab:study4-expSetupScenarioMovements}. 

\begin{table}[!t]
\centering
\captionsetup{font=scriptsize}
\caption{Description of parameters to transform EPA ratings into corresponding LED attributes. We determined the range of LED parameters based on the feedback through previous pilot studies.}
\label{tab:study4-EPAvsLightsConversion}
\scriptsize
\begin{tabular}{c|l|l|l|l}
                & & \multicolumn{3}{|c|}{\textbf{LED Parameters}}                     \\
               \cline{3-5}
\textbf{Name}  & \textbf{Description}  & \textbf{Description} & \textbf{Min} & \textbf{Max} \\
\toprule
Evaluation (E) & Goodness   & Color                & Red & Green        \\
Potency (P)    & Powerfulness  & Intensity            & 0                          & 255          \\
Activity (A)   & Activeness  & Duration            & 4300 ms & 300 ms \\
\bottomrule
\end{tabular}
\end{table}

\textbf{Step 3:} participants were asked questions regarding their opinions about search and rescue robots, emotions in SAR, and how difficult they found the noise levels of the displayed text. These questions asked: (1) how useful they thought rescue robots are, (2) how familiar they were with rescue robots, (3) whether they had seen a SAR robot before, (4) how necessary they thought rescue robots are, (5) how much they believed rescue robots could be better than rescue dogs in the future\footnote{Note, in our research we do not intend to suggest that robots might be better than rescue dogs (which probably they cannot, in many ways), but included this question to provide participants with a more familiar reference.}, (along with 3 additional consistency and sanity checks).  Next, participants answered a question depending on the condition they were assigned. Participants in the emotion condition were asked to report on (9) how much they thought the robot's use of lights in order to convey emotions was helpful to understand messages sent by the robot, while the participants in the no emotion condition had to state (9) if they preferred the robot to use lights/emotions and thought that could be helpful for understanding the messages sent by the robot. All questions in this section were rated on a continuous scale, and participants had an option of ``prefer not to share'' if they did not wish to provide answers (as recommended by the ethics committee).%

\subsection{Apparatus and Simulations}

\subsubsection{Husky Robot}
Husky is an Unmanned Ground Vehicle (UGV) designed by Clearpath Robotics\footnote{https://clearpathrobotics.com/husky-unmanned-ground-vehicle-robot/} to be used as outdoor field research robot. It fully supports Robot Operating System (ROS).

\subsubsection{Emotional Expressions Using Lights}
\label{section:study4-lightimplementation}

A NeoPixel RGB LED strip was used. All emotions that were used in the experiment were programmed on an Arduino micro-controller in C++ using libraries Adafruit NeoPixel and FastLED. Each emotion was given a function in which the period, wavelength, and color of the wave could be altered based on the design. For the experiment, EPA dimensions were transformed to represent different parameters of LED lights (see Table~\ref{tab:study4-EPAvsLightsConversion} for description of parameters and  Table~\ref{tab:study4-EPAvsLightsConversion2} for parameter values). Related software will be made open-source \textit{upon acceptance of the article} to provide a starting point for other researchers who are interested in implementing affective expressions based on ACT.\footnote{\url{www.github.com/samialperen/epaLights}}

For recording the videos of Husky showing affective expressions using LEDs, we attached two LED strips to the Husky robot's top and side (360\si{\degree}) to provide better perception of light from various various viewing angles, e.g.  Figure~\ref{fig:study4-mainTaskBothVideosandQuestion}. We normalized the range of LED parameter values based on the EPA range of emotions under consideration. For example, emotion ``happy'' has the largest P value (2.85), so it was converted to 255 (max LED value for RGB parameters). We took the risk of a seizure into account while selecting the minimum duration~\cite{wilkins2010led}.

\begin{table}[!t]
\centering
\captionsetup{font=scriptsize}
\caption{Values of parameters to transform EPA ratings (min -4.3 and max 4.3) into corresponding LED attributes for each emotion used in the study. EPA values for the emotions were obtained from the USA Student 2015 dictionary~\cite{smith2016mean}.}
\label{tab:study4-EPAvsLightsConversion2}
\footnotesize
\begin{tabular}{c|c|c|c|c|c|c|c}
               & \multicolumn{3}{c|}{\textbf{EPA Values~\cite{smith2016mean}}}              & \multicolumn{4}{c}{\textbf{LED Parameter Values}}                     \\
               \cline{2-8}
\textbf{Emotion}  & \textbf{E}            & \textbf{P} & \textbf{A} & \textbf{R} & \textbf{G} & \textbf{B} & \textbf{Duration(ms)} \\
\toprule
tired & -1.58          & -1.28         & -2.28          & 31                & 0 & 0 & 4154        \\
happy & 3.54          & 2.85         & 0.85          & 0                & 255 & 0 & 1609        \\
fear & -2.41          & -1.07         & -0.81          & 54                & 0 & 0 & 2958        \\
excited & 2.77          & 2.13         & 2.46          & 0                & 174 & 0 & 300        \\
annoyed & -2.13          & -0.47         & 0.58          & 64                & 0 & 0 & 1828        \\
\bottomrule
\end{tabular}
\end{table}

\subsubsection{Gazebo Simulation}
Gazebo simulator was used with ROS to create a realistic SAR simulation~\cite{quigley2009ros}. To construct the SAR disaster environment, various 3D models provided by Open Robotics\footnote{\url{https://github.com/osrf/gazebo_models}} were combined based on experimental procedure.The resulting simulation environment in Gazebo will be made publicly available upon acceptance of this article so that other researchers could use it for their own research.

\subsection{Participants}
We recruited 151 participants on Amazon Mechanical Turk. Only participants whose approval rate was higher than 97\% based on at least 100 HITS were allowed to join the study to increase the quality of the obtained data. Recruited participants were located either in the USA or in Canada. Participants who completed the study were paid \$3 for compensation, while a pro-rated amount was paid to those who did not finish the task. This study received full ethics clearance from the University of Waterloo's Human Research Ethics Board.

16 of the participants failed an attention check question (i.e., ``I think drinking water is liquid''). Also, data of 33 participants who gave inconsistent responses were discarded. After filtering, data from 102 participants (37 female, 65 male; ages 22-69, avg: 38.9, std: 11.1) were  left for the analysis where 53 were in the emotion condition and 49 were in the no emotion condition.

\subsection{Statistical Analysis}
In this experiment, we investigated two factors: perception accuracy and response time. Perception accuracy was investigated to address our fourth research question (RQ4). Response time provided us with information on how fast participants responded to the questions (to see if showing emotions affected response time, and also as a way to check how response time differed for those who passed and failed the emotion training, which could provide some insight on whether failure in training was due to participants' level of attention to the task or other factors).

The perception accuracy of participants was calculated by measuring their success in selecting the correct SAR-related messages. Response time was reported by measuring how fast they selected the messages. 

Further, the independent measures considered in this study were: (a) participants' responses to the questions in the survey, (b) the order of messages seen by the participants, (c) the total number of times they switched from the main task, (d) the total inactive time not spent on the task, and (e) the experimental condition participants were assigned to (emotion vs. no emotion). To investigate the relation between the independent and dependent measures, Linear Mixed Effect Models (LMMs)~\cite{bates2005fitting} were employed and the factors in the model were decided based on minimizing Akaike’s Information Criterion (AIC)~\cite{arnold2010uninformative}. One-way binomial tests were applied assuming uniform probability distribution as the null hypothesis to determine whether participants selected a specific scenario (or emotion in training step) significantly more than another option~\cite{saldien2010expressing}.

\subsection{Results}
Through LMM, it was found that participants in the emotion condition had a significantly higher perception accuracy than participants in no emotion condition ($se=0.04, t=2.287, p=.024$). On the other hand, no significant correlation was found between the condition participants were assigned to and their response time ($se=6.06, t=-0.05, p=.960$). These results are shown in Tables~\ref{tab:study4-lmmTableAccuracy} and~\ref{tab:study4-lmmTableRespTime}. 

\begin{table}[!b]
\centering
\captionsetup{font=scriptsize}
\caption{Linear Mixed-effects predicting participants' perception accuracy}
\label{tab:study4-lmmTableAccuracy}
\scriptsize
\begin{tabular}{lllll}
\hline
\textbf{Covariate}             & \multicolumn{4}{c}{\textbf{Perception Accuracy}}      \\
\cline{2-5}
                               & Estimate  & SE       & t      & Pr ($> |t|$) \\
\hline
\textbf{Condition}             &           &          &        &                       \\
No Emotion~\textsuperscript{b}                   &           &          &        &                       \\
Emotion                        & 0.09      & 0.04     & 2.287  & 0.024 *               \\
\hline
\textbf{Familiar with SAR}     & -1.64e-04 & 6.78e-05 & -2.417 & 0.017 *               \\
\hline
\textbf{SAR Robots Not Useful} & -2.59e-04 & 8.02e-05 & -3.234 & 0.002 ** \\ 
\hline
\end{tabular}
    \begin{tablenotes}
      \centering
      \vspace{1px}
      \scriptsize
      \item * = $p < .05$; ** = $p < .01$; *** = $p < .001$; \textsuperscript{b} = baseline level
    \end{tablenotes}
\end{table}

\begin{table}[!b]
\centering
\captionsetup{font=scriptsize}
\caption{Linear Mixed-effects prediction participants' response time}
\label{tab:study4-lmmTableRespTime}
\scriptsize
\begin{tabular}{lllll}
\hline
\textbf{Covariate}             & \multicolumn{4}{c}{\textbf{Response Time}}     \\
\cline{2-5}
                               & Estimate & SE   & t      & Pr ($> |t|$) \\
\hline
\textbf{Condition}             &          &      &        &                       \\
No Emotion~\textsuperscript{b}                   &          &      &        &                       \\
Emotion                        & -0.3     & 6.06 & -0.05  & 0.960                  \\
\hline
\textbf{Inactive Time}         & 0.11     & 0.03 & 3.273  & 0.001 **              \\
\hline
\textbf{SAR Robots Useful}     & 0.04     & 0.02 & 2.012  & 0.047 *               \\
\hline
\textbf{Familiar with SAR}     & 0.06     & 0.01 & 4.355  & 0.000 ***   \\
\hline
\textbf{Seen SAR Robot Before} & -0.04    & 0.01 & -2.865 & 0.005 **              \\
\hline
\textbf{SAR Robots Necessary}  & -0.03    & 0.02 & -1.543 & 0.126                 \\
\hline
\textbf{Order Number}          & -6.1     & 2.45 & -2.487 & 0.013 * \\ 
\hline
\end{tabular}
    \begin{tablenotes}
      \centering
      \vspace{1px}
      \scriptsize
      \item * = $p < .05$; ** = $p < .01$; *** = $p < .001$; \textsuperscript{b} = baseline level
    \end{tablenotes}
\end{table}

LMM results also revealed other significant findings. As suggested in Table~\ref{tab:study4-lmmTableAccuracy}, participants who thought rescue robots were not useful or who were familiar with SAR had a significantly lower perception accuracy. Furthermore, Table~\ref{tab:study4-lmmTableRespTime} shows that while there was no significant differences between the conditions, participants familiar with SAR, participants who think rescue robots are useful, or those who had a higher inactive time spent more time predicting the messages in the scenarios. On the other hand, participants who had seen a rescue robot before were faster to respond. We also detected that participants got faster in providing the response as they see more videos (shown as order effect in the table).

\subsubsection{Training Success}

Training success is important specifically in the emotion condition, as emotions might not have provided a beneficial additional communication modality for those who did not pass the training. 
\begin{table}[!t]
    \centering
    \captionsetup{font=scriptsize}
    \footnotesize
    \caption{The table shows participants' incorrect guesses to recognize implemented affective expressions on Husky. Rows show emotions expressed by Husky, and columns show participants' corresponding responses to these expressions. Since only incorrect responses of participants who failed to learn all emotions in the training step are included (41 participants failed in total; 22 in emotion condition, 19 in no emotion condition), all diagonal entries have zero, i.e., a non-zero diagonal entry means that the emotion was  perceived correctly.}
    \label{tab:study4-emotionTraining}
    \scriptsize
    \begin{tabular}{c|c|c|c|c|c|c}
      & \bf \rotatebox[origin=c]{0}{Responses}
      & \bf \rotatebox[origin=c]{0}{Tired} & \bf \rotatebox[origin=c]{0}{Fear} & \bf \rotatebox[origin=c]{0}{Excited} & \bf \rotatebox[origin=c]{0}{Happy} & \bf \rotatebox[origin=c]{0}{Annoyed} 
      \\ \toprule

        \multirow{5}{*}{\rotatebox[origin=c]{90}{\textbf{Emotions}}} & Tired & 0 & 7 & 4 & 4 & 10 \\ 
        \cline{2-7}
        &Fear & 26 & 0  & 4 & 4 & 14 \\ 
        \cline{2-7}
        &Excited & 0 & 1 & 0 & 20 & 1 \\ 
        \cline{2-7}
        &Happy &  2 & 1 & 5 & 0& 1 \\ 
        \cline{2-7}
        &Annoyed & 14 & 27 & 7 & 4 & 0 \\ 
        \bottomrule
     \end{tabular}
\end{table}

41 participants failed to learn all five emotions during training. They were labeled as ``failed'' to investigate their results separately. Table~\ref{tab:study4-emotionTraining} shows only incorrect responses of these 41 participants. Among the mis-recognized emotion pairs, fear-tired and annoyed-fear were the ones that got confused the most, while the happy-excited pair was the least confused. All emotions were perceived with an accuracy more than the chance (20\%), while ``happy'' was perceived with the best accuracy ($\approx91\%$), and ``annoyed'' with the worst accuracy ($\approx59\%$).

\subsubsection{Perception Accuracy - Training Success}

\begin{figure} [htbp]
    \centering
    \captionsetup{font=scriptsize}
    \subfloat[]{
    \includegraphics[width=0.48\columnwidth]{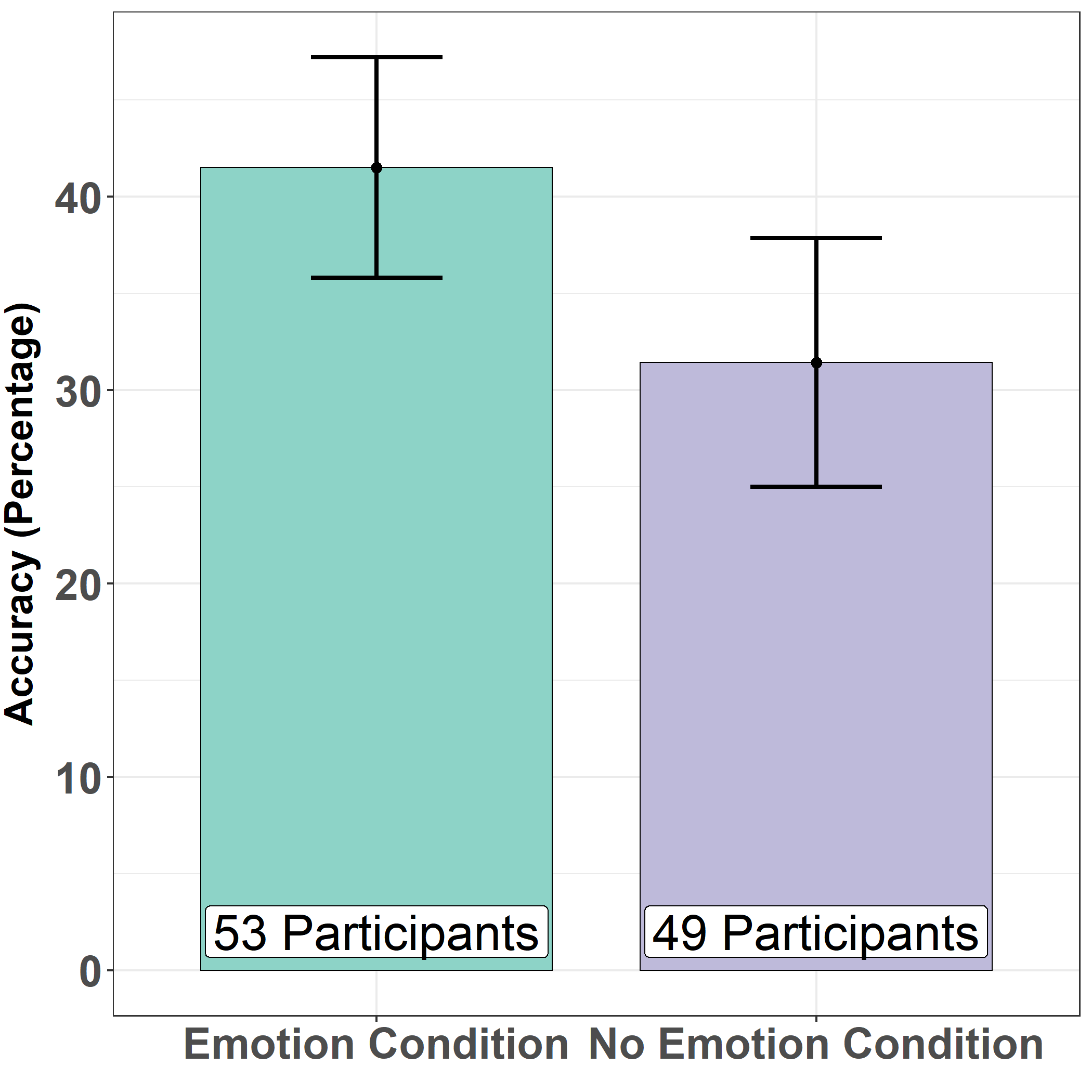}}
    \hfill
    \subfloat[]{ \includegraphics[width=0.48\columnwidth]{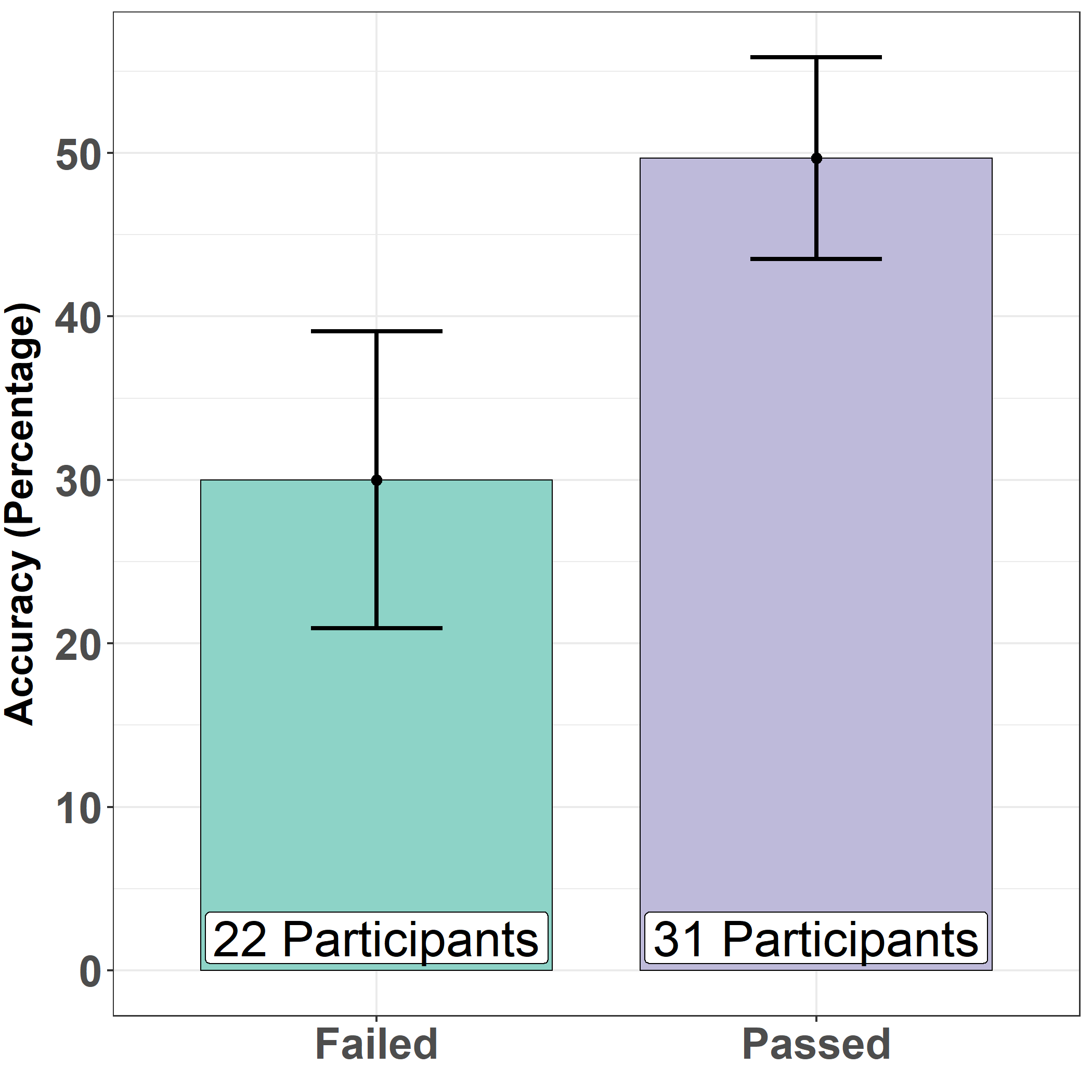}}
    \caption{(a) Average perception accuracy of participants for each condition given (difference is significant; $t=2.357, df = 97.623, p=0.020$). (b) Average perception accuracy of participants in the emotion condition compared based on their success during the emotion training step (difference is significant; $t=2.425, se=0.054, p=0.019$).}
    \label{fig:study4-avgPerceptionAccuracy}
\end{figure}

As mentioned before, the participants in the emotion condition had a significantly higher accuracy than participants in the no emotion condition. The average perception accuracies of participants in both conditions is shown in Figure~\ref{fig:study4-avgPerceptionAccuracy}. 
Participants in the emotion condition were also divided into two groups  depending on their success during the training. Those who passed the training step had a significantly higher accuracy than those who failed based on LMM (t=2.425, se=0.054, p=0.019). We did not find a significant difference between perception accuracies of participants who passed and failed attention checks in the no emotion condition. 

\begin{figure*}[!ht]
    \centering
    \captionsetup[subfloat]{font=scriptsize}
    \captionsetup{font=scriptsize}
    \subfloat[Emotion condition - passed the training]{\includegraphics[width=0.69\columnwidth]{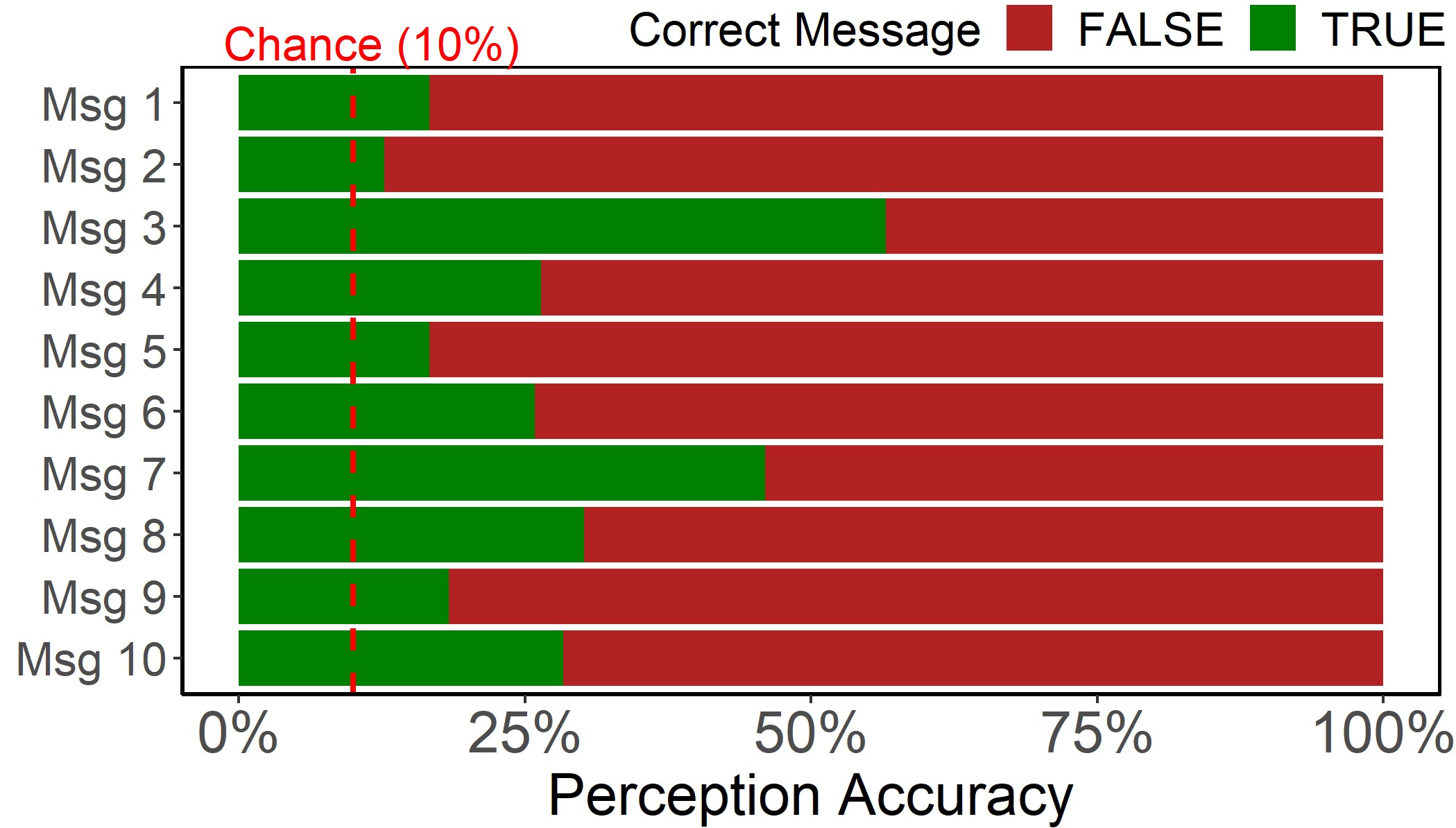}}
    \subfloat[Emotion condition - failed the training]{\includegraphics[width=0.69\columnwidth]{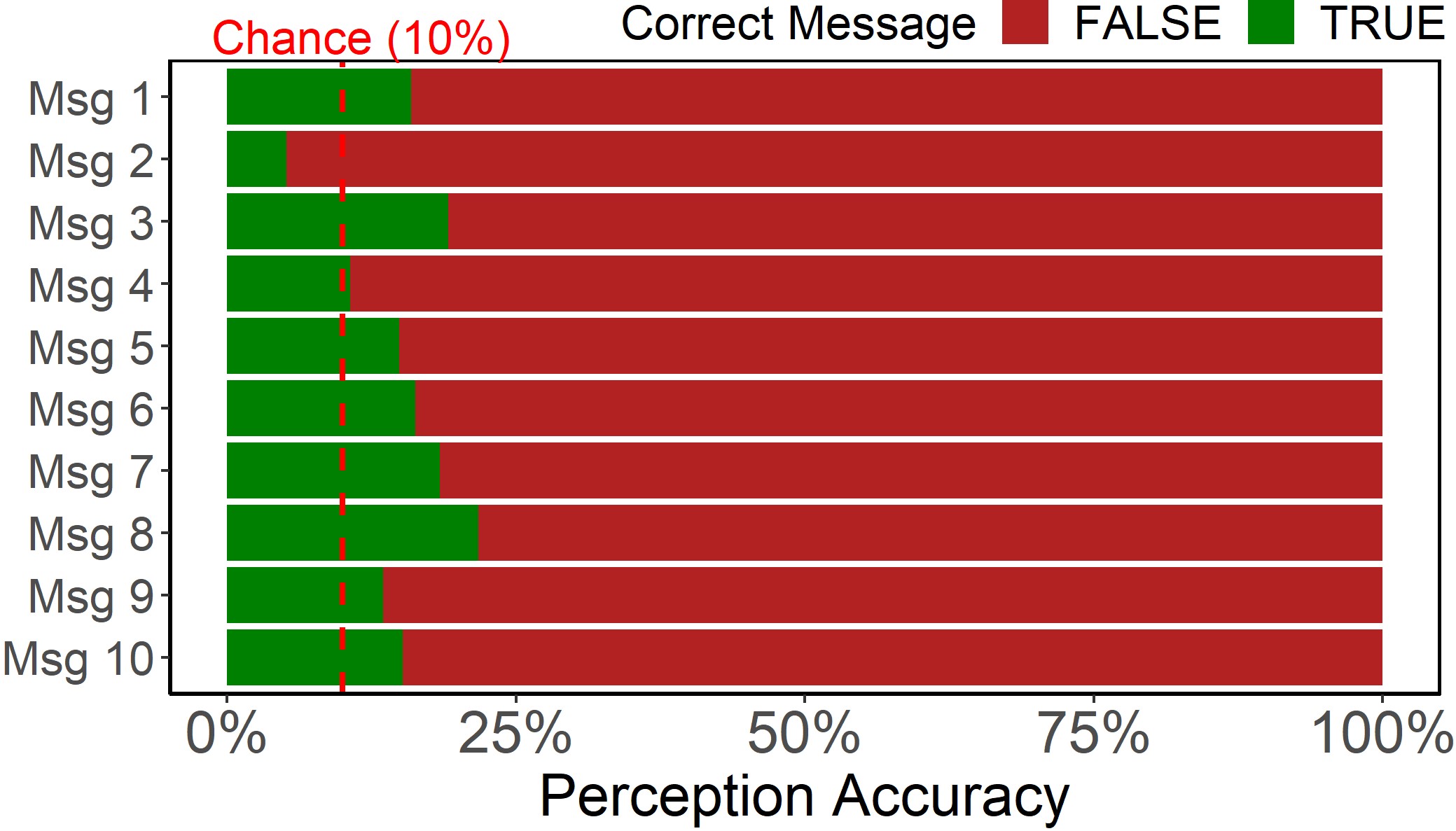}}
    \subfloat[No emotion condition]{\includegraphics[width=0.69\columnwidth]{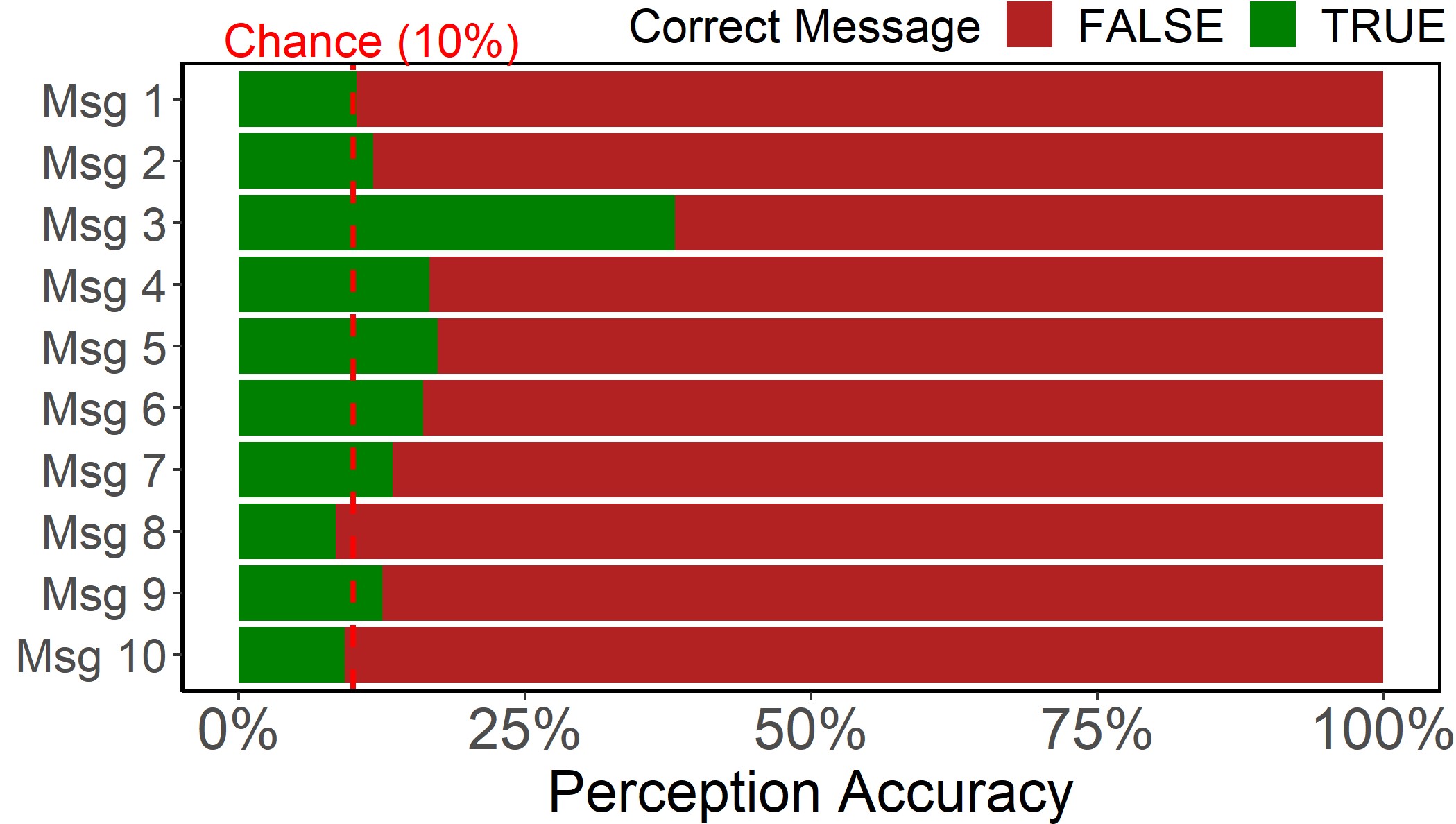}} \\
    \subfloat[Emotion condition - passed the training]{\includegraphics[width=0.69\columnwidth]{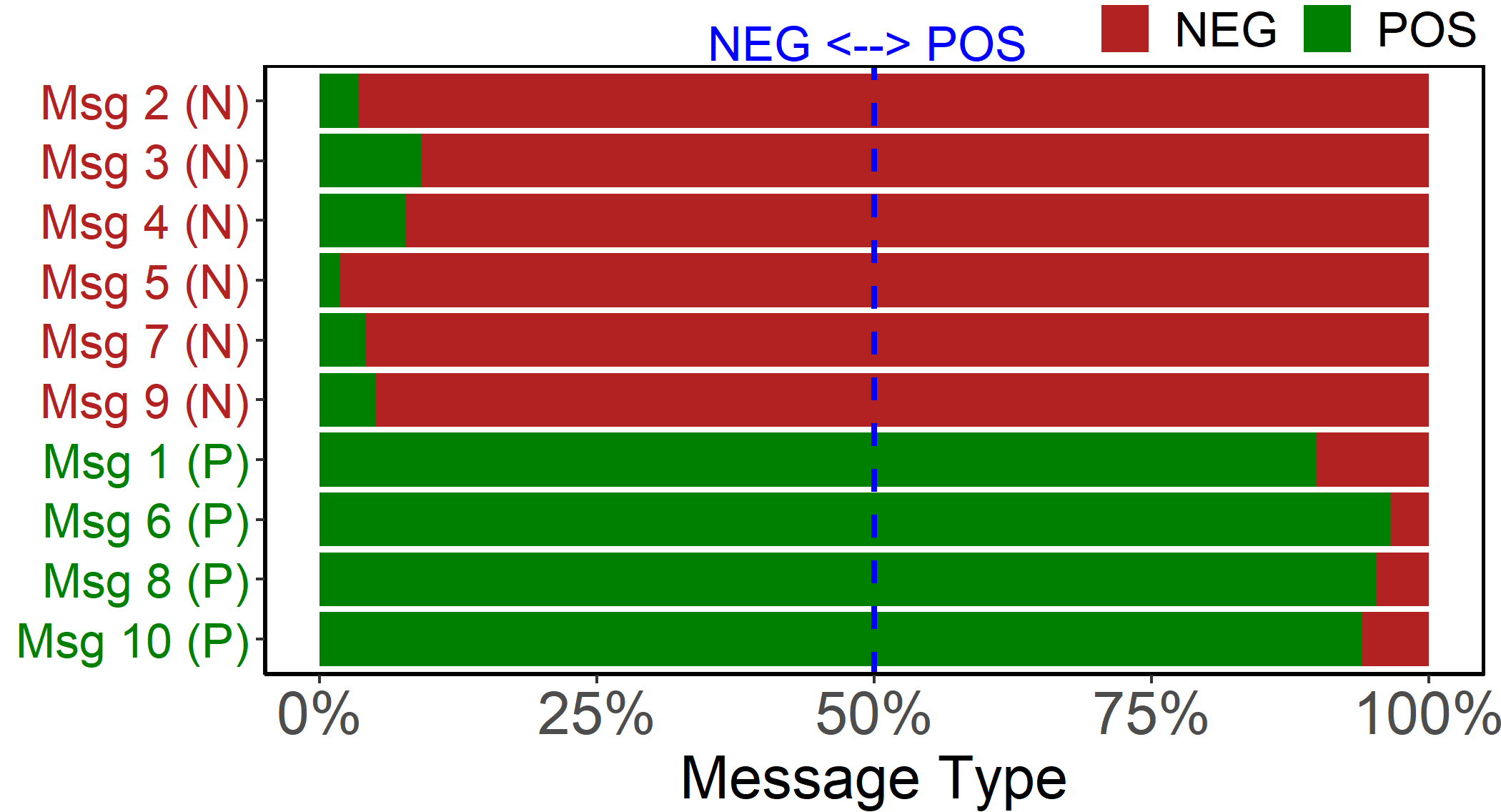}}
    \subfloat[Emotion condition - failed the training]{\includegraphics[width=0.69\columnwidth]{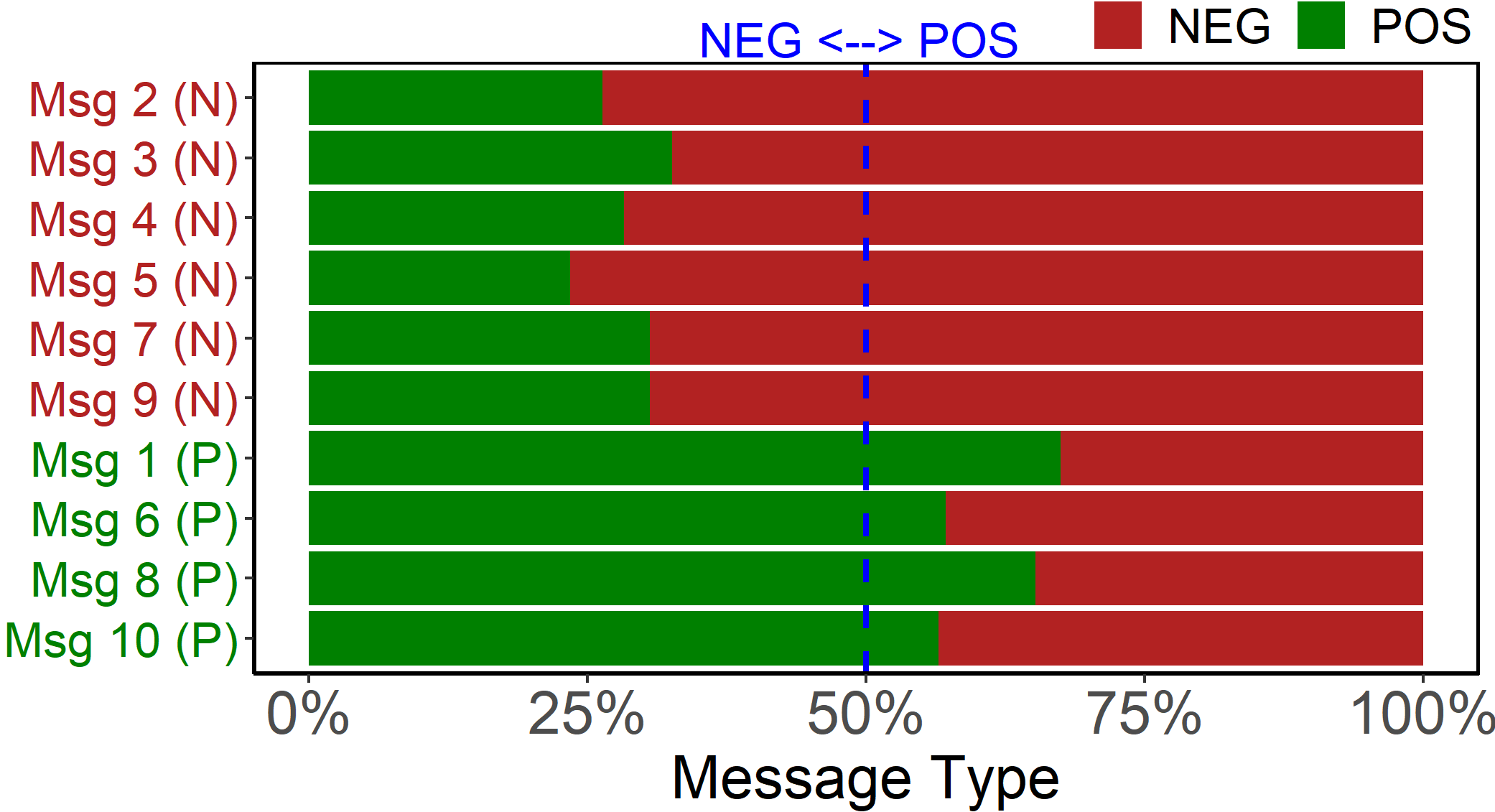}}
    \subfloat[No emotion condition]{\includegraphics[width=0.69\columnwidth]{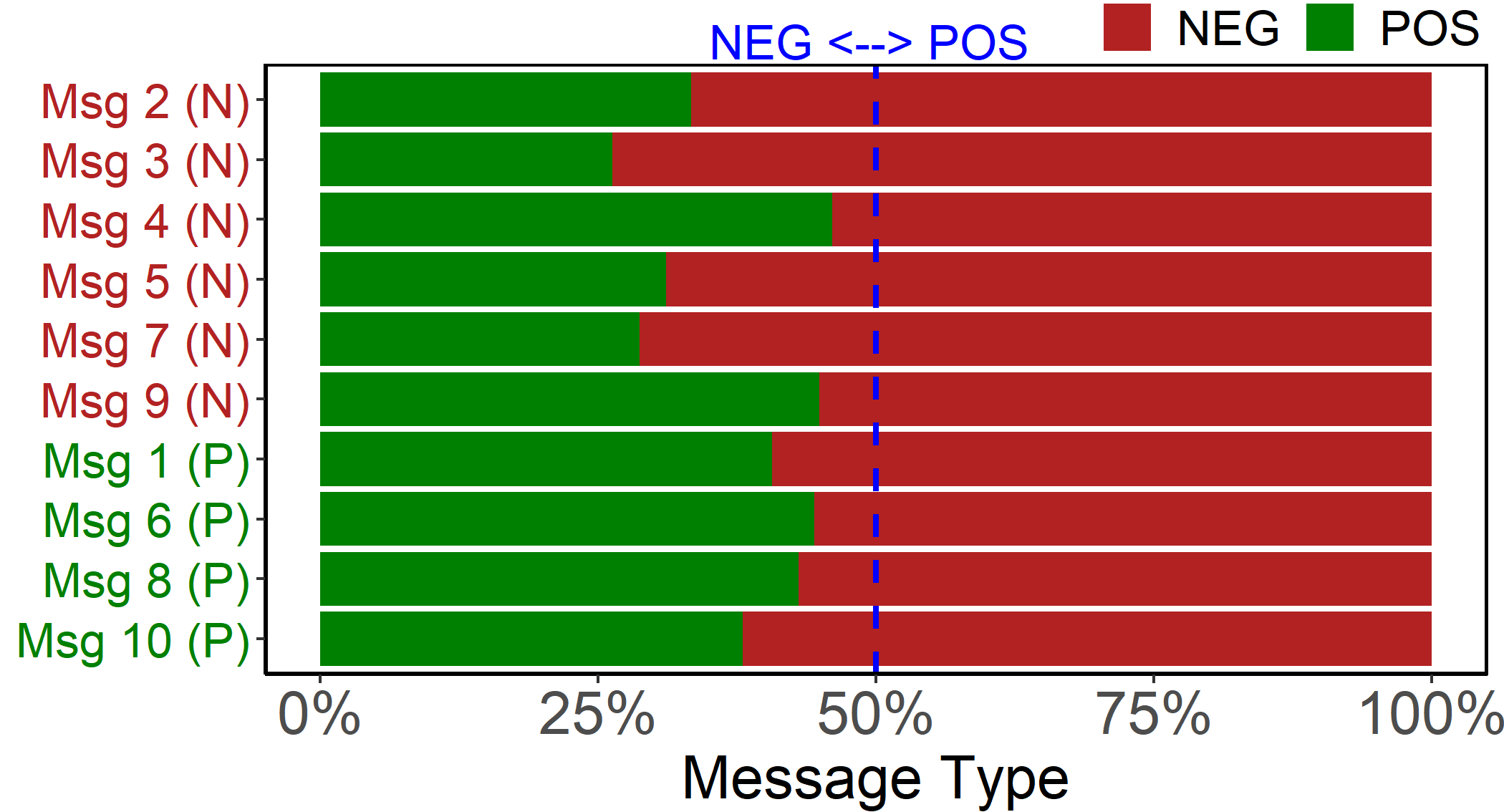}} 
    \caption{Perception accuracy of the messages shown to participants during the experiment 3. Individual accuracies of the messages are shown at the top and accuracies regarding the type of the messages (positivity/negativity) shown at the bottom. See Table~\ref{tab:study4-expSetupScenarioMovements} for the scenario names.}
    \label{fig:study4-scenarioRecognitionBarPlots}
\end{figure*}

The success of recognizing individual SAR scenarios in terms of robot to human communications was also analyzed for three groups: (a) participants in the emotion condition who passed the training, (b) participants in the emotion condition who failed the training,  and (c) participants in the no emotion condition including those who failed and passed the training step (see Figure~\ref{fig:study4-scenarioRecognitionBarPlots}). Overall, those who were assigned to the emotion condition and passed the emotion training had a significantly higher accuracy as compared with both those who failed in the emotion condition (se=0.057, t=-2.273, p=0.025), and those in the no emotion condition (se=0.045, t=-3.211, p=0.002), according to a LMM. Note, none of the scenarios were recognized with more than 60\% accuracy.

The success in recognizing individual SAR scenarios regarding their type (positive vs. negative sentiment, based on the Evaluation dimension of the existing EPA values) was also examined for the same three participant groups (see Figure~\ref{fig:study4-scenarioRecognitionBarPlots}). Participants who passed the training in the emotion condition, shown in Figure~\ref{fig:study4-scenarioRecognitionBarPlots}~(a), had the highest accuracy in understanding whether the scenario was positive or negative (over 90\% accuracy). In contrast, participants in the no emotion condition, shown in Figure~\ref{fig:study4-scenarioRecognitionBarPlots}(c), had the lowest accuracy (for some scenarios, their accuracy was even less than the chance level, i.e., 50\%).

\subsubsection{Questionnaire Results}
\begin{figure}[tb]
    \centering
    \captionsetup{font=scriptsize}
    \captionsetup[subfloat]{labelformat=empty}   
    \subfloat[]{\includegraphics[width=0.49\columnwidth]{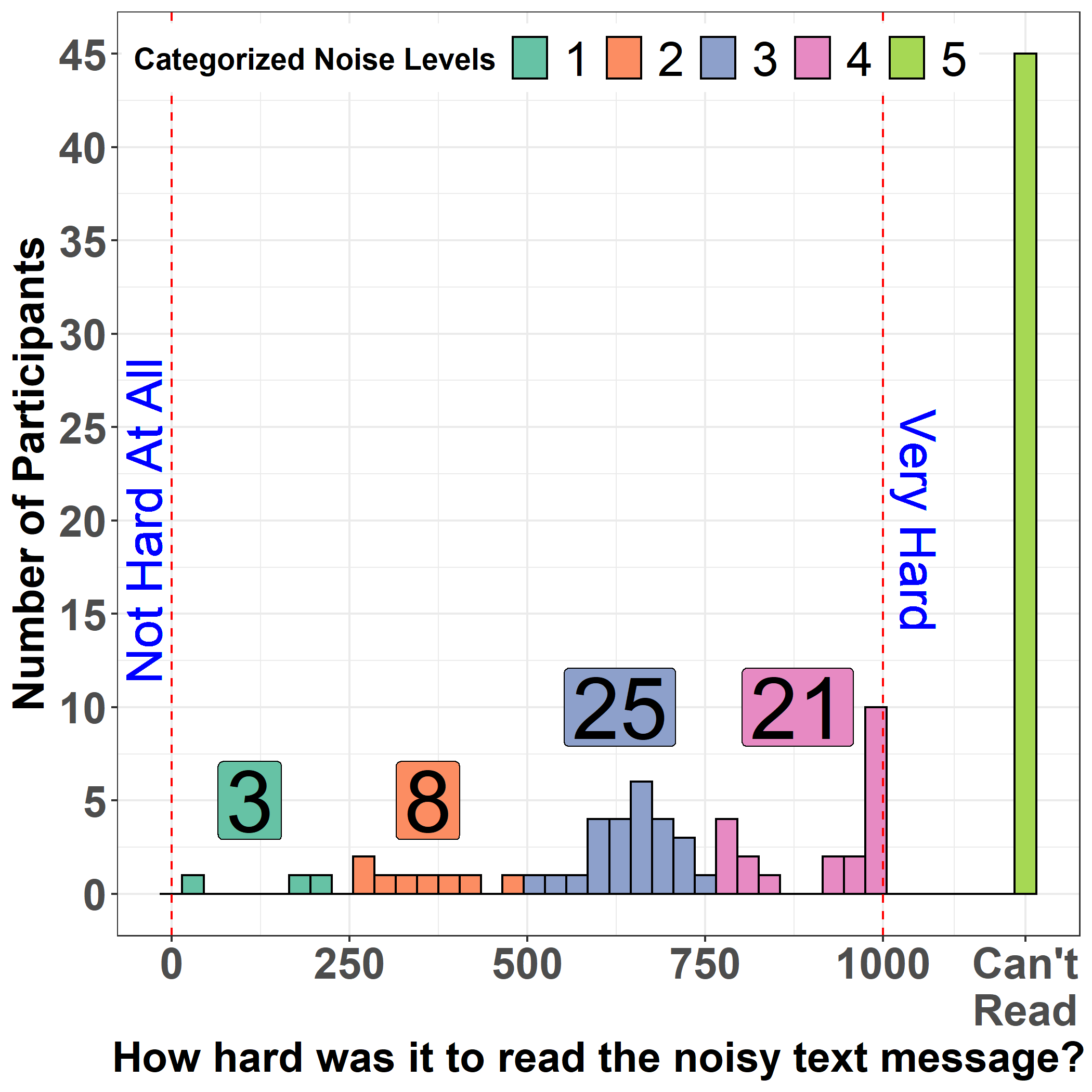}} 
    \subfloat[]{\includegraphics[width=0.49\columnwidth]{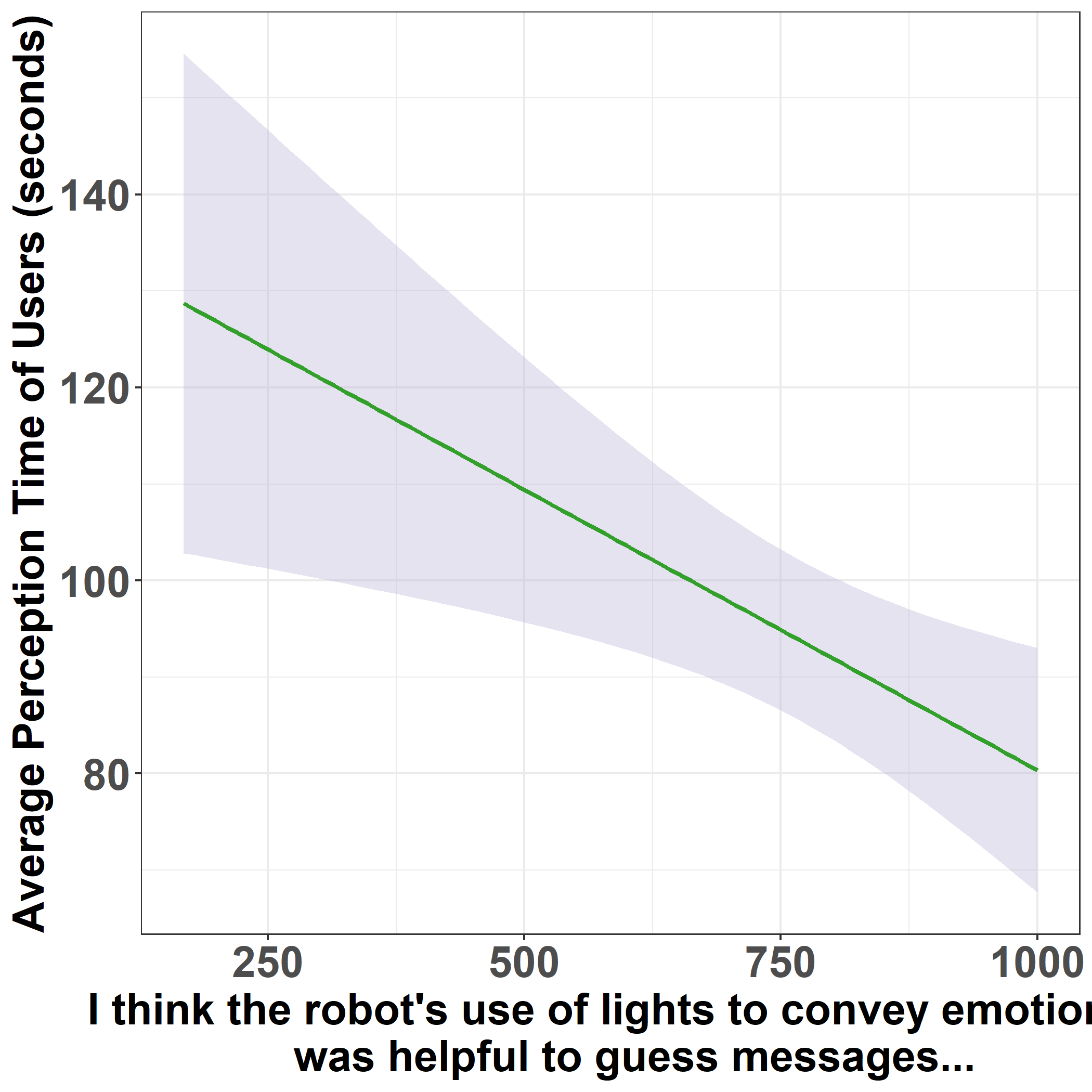}}
    \caption{On the left: Participants' responses to the survey question ``How hard was it to read the noisy text messages?'' and on the right: average response time of each participant vs how useful they found lights to guess messages sent by the robot}
    \label{fig:study4-hardtextDistribution}
\end{figure}

Participants in both conditions were asked to report how hard it was for them to read the distorted text messages. As it can be seen in Figure~\ref{fig:study4-hardtextDistribution}, the majority of the participants found the shown text messages very difficult to read, with 45 of them stating that they could not read the text. To test the relation between their responses to this question and their performance during the main task, reported noise levels were factorized into five, with noise level 1 representing their responses between 0 and 250, and level 5 representing that they could not read the text at all. We did not observe an effect of the reported noise level on perception accuracy~($se = 0.018, t = -0.520, p = 0.604$) \& response time ($se = 2.818, t = -1.204, p = 0.231$).

\begin{figure}[htb]
    \centering
    \captionsetup{font=scriptsize}
    \subfloat[No Emotion Condition]{\includegraphics[width=0.49\columnwidth]{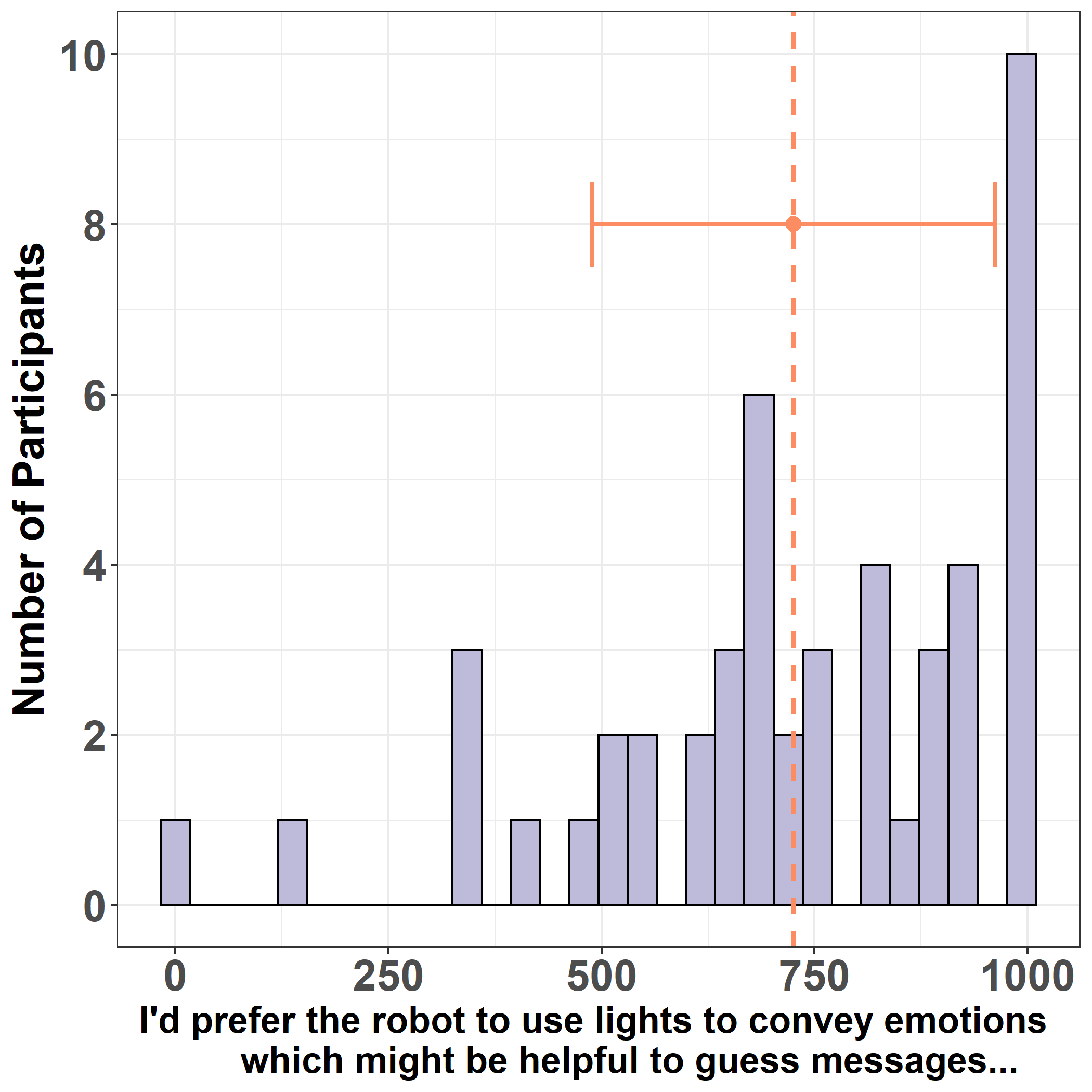}} 
    \subfloat[Emotion Condition]{\includegraphics[width=0.49\columnwidth]{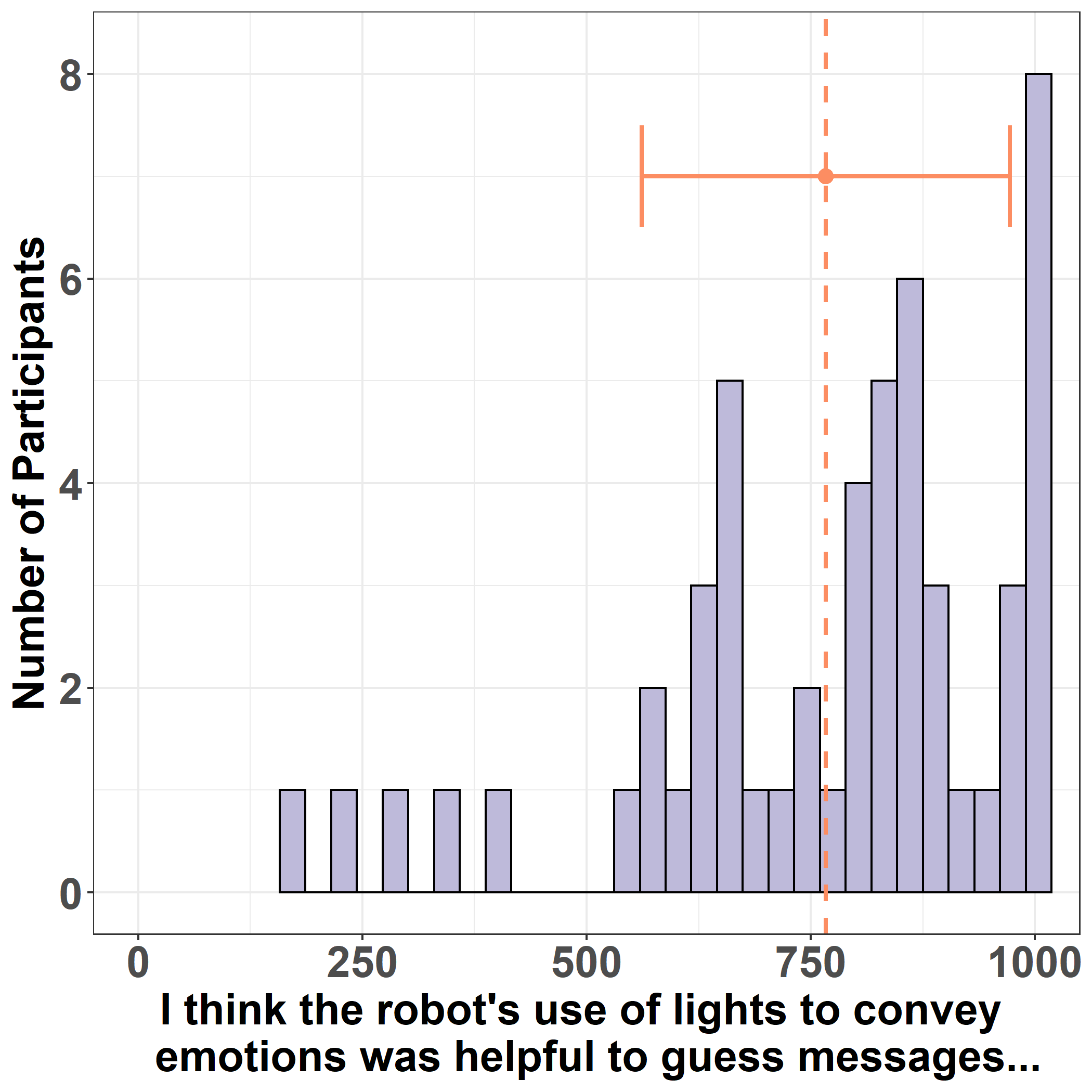}}
    \caption{Response of participants in both conditions. Mean and standard deviation were shown with orange dashed and solid lines respectively.}
    \label{fig:study4-helpfulPreferLightsDistribution}
\end{figure}

Participants' responses to the two statements about the usage of affective lights during the experiment are shown in Figure~\ref{fig:study4-helpfulPreferLightsDistribution}.  For both statements (one for each condition), the mean value is around 750 (1000 corresponding to `I totally agree' and 0 indicating `I totally disagree'), showing that most of the participants in the no emotion condition indicated that they preferred to see the emotions, and those in the emotion condition found the robot's emotions to be helpful in understanding the situation.

A linear mixed-effect model was fit for participants in the emotion condition to predict response time and perception accuracy based on their ratings of how helpful they found the lights. While there was no significant effect of ratings on perception accuracy ($se = 0.000, t = 0.819, p = 0.413$), there was a significant negative effect of how helpful they found the lights on their response time ($se = 0.000, t = -2.479, p = 0.016$), indicating that the more useful the participants rated the emotions, the less time they spent on guessing the messages (see the plot on right in Figure~\ref{fig:study4-hardtextDistribution}).

\section{Discussion}

In this paper we studied the feasibility of using emotions as a communication modality in SAR robots. Through three online experiments, considering a range of different scenarios that could occur in search and rescue situations, we provided evidence that emotions might in fact be useful as an additional communication channel in SAR robots -- to complement the existing communication modalities and to improve the success and efficiency of robot to human communication. As an additional benefit, this ability might also help victims since it has been suggested that social SAR robots can contribute to the reduction of stress levels of victims and prevent shock~\cite{Bethel2011}. However, as using emotions in SAR robots for communication purposes is a novel research direction, we carried out three experiments into understanding (a) the feasibility of using emotions in SAR robots (Experiment~1), (b) different approaches that can be used to decide on an emotion that a SAR robot should show in a specific situation for communicating a specific message, and (c) whether emotions can improve communications between SAR robots and humans when other modalities fail.

In Experiment~1, we asked whether there would be consensus in mapping emotions to SAR situations (\textbf{RQ1}), and if such a mapping would be robust and not be affected by the wording of sentences in those situations (\textbf{RQ2}). The results showed that participants agree on a particular mapping between described SAR situations and emotions of a SAR robot to convey that situation, which was not affected by the wording style. These results were promising and encouraged us to further pursue the use of  emotions as an additional communication channel for SAR robots. 

As these mappings might be affected by the selected set of emotions (e.g., a specific SAR robot may not be capable of showing specific emotions), we asked if it is possible to use a method to get the mappings in a way that would be flexible and independent of the selected set of emotions, to address \textbf{RQ3}. This led to the design of Experiment~2, where Affect Control Theory (ACT) was used and mappings were measured along three different dimensions: Evaluation, Activity, and Potency (EPA). We then used the EPA values associated with the set of 11 emotions (used in Experiment~1) to check whether the mappings would be consistent between the two experiments. Results suggested that similar mappings can be obtained when participants are asked to rate SAR situations on the EPA dimensions, as opposed to directly mapping to emotion words. Therefore,  mappings using EPA ratings can be used in the future, which can flexibly be used with different emotion sets.

Dimensional emotion models are not usually used to understand emotions of sentences. They are rather often used to describe emotions (as well as identities and behaviours in ACT). That is because it is difficult to obtain ratings for these dimensions for sentences due to multiple challenges. For example, the mappings will be highly context-dependent, and it would be hard (if not impossible) to conduct extensive surveys to gather ratings for all combinations of sentences, in all contexts, in a similar manner that the other EPA values are collected (as a large number of sentences can be created with the combination of the related words). Further, there is currently limited literature on mapping sentences to emotions or dimensional emotion values, most of which can only evaluate sentences on the Evaluation dimension (i.e., regarding a sentence's sentiment). Since automated methods for inferring emotions from sentences are still not reliable, in Experiment~2 we presented a set of context-dependent (i.e., SAR related) EPA values. Note, different robots have different capabilities and limitations regarding affective expressiveness. However, with the proposed method we developed in Experiment~2, depending on the robot's expressive capabilities (e.g. in case it is not able to display certain emotions), the set of emotions can be adjusted by identifying the ``next best" emotions. Thus, the methods allows us to find mappings according to the robots' capabilities. However, it is important to emphasize that, regardless of the robots' capabilities, we still need to decide on a reasonable set of emotions that are being considered for the mappings. The ACT datasets include a large set of emotions, many of which may not be relevant in some specific contexts. For example, if we had used the complete set instead of limiting it to our set of 11 emotions, the two closest affective expressions for the sentence ``I detected dangerous material here, let's proceed carefully" would have been ``obligated" and ``aggrieved", which may (a) not be appropriate for the context of SAR, and (b) not be feasible to express on a social robot. A similar approach might be used in other application domains beyond search and rescue, however, future work is needed to study generalizability of our approach to other domains. Also, as emotions are context dependent~\cite{cohen2014context,ma2018socio}, while a similar method could be used in other domains to obtain mappings between situations/messages and emotional displays of a robot, it is reasonable to expect that the mappings might not be exactly the same.

An interesting finding in Experiment~2 was that while the majority of the sentence ratings from Canada and the USA were highly correlated, for three sentences describing SAR situations, these ratings were either not or only weakly correlated (See Table \ref{tab:low_correlations}). One possible explanation might be that the cultural differences were reflected in these situations more than the others. For example, participants in the USA rated the sentence ``I think we need additional team members" as \textit{better}, \textbf{more powerful}, and \textbf{more active}, as compared to the participants in Canada, who rated this sentence closer to neutral in all dimensions. We saw a similar tendency for the sentence ``I think we have more team members than we need. One of us should join the other team". In both cases, the situation involved a change in the structure of teams. While future work is needed to study why the ratings were not correlated for these few situations and whether cultural differences were in fact the reason behind this observation, if this is the case, it may emphasize that cultural differences should also be considered while designing emotions for robot to human communication of SAR robots (similar to the way different EPA ratings are obtained in different countries for emotions, identities, and behaviours in ACT). These differences can be important considerations for the future design of robots' emotional displays in different application domains~\cite{ma2018socio}.

As discussed above, the first two studies supported the feasibility of bringing emotions into the SAR context. Therefore, in Experiment~3 we investigated the effectiveness of using emotions as an additional communication channel (e.g., video streams, voice, and text~\cite{jones2020}). Our intention is to propose an additional interaction modality to \textit{complement} existing multi-modal channels. In this way, we would be able to employ SAR robots with more robust and failure-safe communication abilities that might help to improve field workers' shared mental model and situational awareness~\cite{seppanen2013developing}. To that end, we first proposed a method to show affective expressions on an appearance constrained robot, Husky, using lights, designed also based on ACT and EPA dimensions. The proposed method was inspired by an earlier work in~\cite{collins2015saying}. Employing EPA dimensions allowed us to implement affective expressions quantitatively on appearance-constrained robots. Using these implemented emotional displays and the mappings obtained from the first two studies, as well as through simulations of SAR scenarios (which were designed in a way to also convey the context that the SAR workers would get from SAR robots' movements, locations, etc.), we investigated whether the usage of emotions can improve the communication in robot-assisted SAR teams (\textbf{RQ4}) in Experiment~3. Results suggested that a rescue robot that uses emotions can increase the accuracy of understanding the messages when other communication modalities (i.e., text in our experiment) fail, supporting \textbf{H1}.
 It is important to emphasize that, in our work, the effect of emotional displays on improving the accuracy of understanding robots' messages was studied in the specific context of search and rescue, and for a range of common situations that often occur in search and rescue. Future work will benefit from studying how emotional displays can complement multi-modal communications between humans and robots in other domains, especially in similar situations where other communication modalities may fail. This includes, but is not limited to, robots operating in noisy environments, or robots interacting with persons who have hearing impairments.

While it was not the focus of this study, we found many challenges with conveying emotions through lights. Even though participants were trained at the beginning about the meaning of each emotional display (a step expected to be part of future training of human team members in future SAR applications), we had participants who failed the training. Fear was the most commonly confused emotion among the negative emotions, which was confused with either annoyed or tired. The reason behind this misrecognition may be that EPA values for our negative emotions were close to each other, which makes them less distinguishable through lights only, especially as captured through videos and displayed on computer screens. Under these condidtions, e.g., the difference in the Potency (P) dimension may not be recognized well solely based on changes in light intensity.  
Therefore, improvement of emotional displays can further improve communications. In our experiment, emotion training accuracy increased participants' accuracy in understanding the SAR situations. Those who successfully passed the emotion training step had a significantly higher accuracy in recognizing the SAR situations. However, even those who failed to distinguish the emotions in the training step could still benefit from the affective expressions by understanding the sentiment (positive or negative Evaluation) of the messages (because sentiment was clearly distinguished by showing green or red lights), leading to less confusion as compared to the no-emotion condition. This suggests that while a larger range of emotions can be helpful in increasing perception accuracy in SAR situations, conveying the sentiment can still be beneficial, limiting the potential SAR messages that a robot may convey.

While it was reasonable to assume that the perceived noise level may also affect accuracy (therefore was controlled for in the analyses), we did not find an effect of the reported noise level on perception accuracy (similar to what was previously seen in~\cite{ghafurian2021recognition}). One explanation could be that the participants judged the difficulty based on how they assumed they performed in reading the text messages (e.g., thinking that they understood the messages while they did in fact not).

To summarize, while future work is needed to better understand the benefits of using affective expressions with SAR robots (e.g., studies in real world situations and recruiting experienced rescue workers), this article provided a first step towards using affective expressions in SAR robots with the goal of increasing efficiency in SAR. We provided evidence on the feasibility and effectiveness of using emotions as an additional communication modality in search and rescue teams, to increase efficiency and robustness of communications, which is a key in success of SAR operations. The idea of using emotions to complement multi-modal human-robot interaction, as well as the proposed methodology for obtaining the mappings and applying them to a specific context, in our case SAR,  has the potential to be applied to other real-world applications that require efficient human-robot teamwork such as in other rescue robots (e.g., firefighting), as long as proper mappings between common situations happening in these contexts and emotions exist.
\section{Limitations \& Future Work}
Our study had several limitations. First of all, due to the online nature of the studies, participants did not have a chance to interact with real SAR robots. They also did not experience a real SAR scenario, which could help with understanding the situations and might affect the mappings. While illustrating possible SAR operations with several pictures of SAR robots as well as simulated and real videos of the Husky robot, the obtained results might differ in real-life scenarios. Mapping are also likely to differ if they were obtained from participants who had experience with SAR situations. This online approach was followed as the first step for this direction of research due to the COVID-19 pandemic. Moreover, the online approach eliminated the change of biasing participants with the appearance of a particular robot, and it also helped with reducing the experimenter bias for the first and second studies~\cite{goodman2013data}. Furthermore, this approach has been shown to be effective in many HCI and HRI studies and has gained more attention since COVID-19 has affected the feasibility of conducting in-person HRI studies, as a safe method for data collection~\cite{10.1145/3405450}. Nonetheless, future work is needed to investigate if/how obtained mappings would translate to real-life situations and with ratings by participants who have experience with SAR situations. Regarding the last experiment, in-person interaction of rescue workers with Husky showing affective expressions should also be investigated in a field study, e.g. simulating a real disaster area.

Although participation was limited to the USA and Canada, participants' level of English was not assessed during any of the studies. Yet, based on their answers to attention check questions, it is reasonable to assume that they understood the task and the sentences.

While Experiment~2 led to ratings that can be used with different emotion sets, we did not examine how the mappings change based on different emotion sets. Future work could obtain these mappings using different emotion subsets, for example those that can be shown by a specific robot. 

In Experiment~3, the first limitation concerns the design of emotional expressions using affective lights. Generally, there are many challenges in designing emotional displays for appearance constraint robots. Emotional expressions have been mostly designed for human-like or zoomorphic robots which are not common in search and rescue scenarios. These challenges could limit the range of emotions that can be shown by appearance constraint robots. In our study, we limited the range of emotions to those that have been previously designed for other types of robots. As an example, we used `calm' as an emotion representing a `near neutral' state, because a neutral state could not be properly designed with lights. Although we do not expect that this has affected the outcome of our study, as all of our messages notified users about an event that was either positive or negative, it could be considered a limitation of our approach. Also, in our study, participants could not distinguish between the different negative emotions as well as they did for the positive emotions. Although positive and negative affective lights' visibly differ in real life, this difference is not that clear in the recorded videos due to the technical difficulties of recording high-speed, low/high brightness of LED lights. An additional study with real human-robot interaction might in fact improve the accuracy of users' emotion recognition. Alternative LED designs could also be investigated to improve the recognition of the robots' emotions shown through lights. Moreover, longer-term, repeated interaction with SAR robots and a better recognition of emotional displays of robots can be expected to improve human-robot communication,  since our findings showed that  participants' success in perceiving SAR scenarios increased as their training success increased (see Figure~\ref{fig:study4-avgPerceptionAccuracy}). Similarly, the selected robot, as well as the specific design of the scenarios, might have affected the findings. Studies that involve SAR robots that are capable of showing a smaller or larger range of emotions, as well as including other scenarios or different designs for these scenarios can complement and verify the findings of the third experiment in future studies.

When affective rescue robots are used in real SAR missions, there might be other challenges regarding the usage of affective expressions in a disaster area. For example, perception of affective expressions might differ in environments with varying visibility conditions such as smoke, rain, or darkness as we investigated in~\cite{ghafurian2021recognition}. While this previous study showed that recognition of affective expressions conveyed through a robot's body and head gestures could be robust, to a reasonable extent, under different visibility constraints~\cite{ghafurian2021recognition}, future work is needed to examine the effects of visibility conditions on the accuracy of recognition of SAR robots' robot-to-human communications through emotions. Also, while we provided a first step towards implementing emotions on appearance-constrained robots, future work needs to investigate how to improve affective expressions of SAR robots using in-person experiments as well as field studies. For example, employing a different way of matching lights parameters with EPA dimensions and/or having additional parameters (like using different patterns for each emotion) can be investigated.

Furthermore, our study was designed in a way so that the majority of the participants could not read the noisy text messages that accompanied the emotional displays. This was because we wanted to first understand the benefits of using emotions as a communication modality in situations when  other modalities fail. Future studies could investigate the impact of emotions in other situations, e.g., when other modalities are less noisy. For example, one could study if using emotions can lead to a faster recognition of the situation. 

Lastly, further studies that employ emotions to convey information from robots to humans in different application areas, such as firefighting and service robotics, can help support generalizability of using emotions as a communication channel to complement multi-modal human-robot interaction in other similar contexts.

\section{Conclusion}
In this article, we presented three online studies to investigate the possibility and benefits of using emotions as a complementary communication modality in robot-assisted Search and Rescue (SAR) to improve communication between rescue workers and SAR robots. Mappings between situations commonly occurring during SAR operations and emotions were obtained through the first Experiment, and a different method was investigated for obtaining the mappings in the second study. Results of the first two studies confirmed the feasibility of using emotions in specific SAR contexts. The mappings were also robust to the wording of the sentences. Employing a dimensional emotion model was investigated and proposed as a practical approach for gathering mappings that are not dependent on a specific emotion set, which could make mappings more generalizable to different SAR robots. In the third Experiment, affective expressions obtained from the mapping obtained in the previous studies were implemented on an appearance-constrained rescue robot (Clearpath Robotics Husky), and were used in simulated SAR situations to study the benefits of using emotions as a communication modality in a situation when other modalities may fail. Results of the third experiment suggested that participants who saw a rescue robot with an ability to express emotions had a better situational awareness. To conclude, this article presented a first step towards using emotions in SAR robots as an additional communication modality and provided insights and approaches that might help with the design and employment of emotions in SAR robots. This is hoped to increase the efficiency of (affective) SAR robots' robot to human communications,  improving participants' situational awareness of the disaster area, which could ultimately  lead to more successful SAR missions.

\ifCLASSOPTIONcompsoc
  \section*{Acknowledgments}
\else
  \section*{Acknowledgment}
\fi
This research was undertaken, in part, thanks to funding from the Canada 150 Research Chairs Program. We would like to thank Hamza Mahdi and Shahed Saleh for their help with designing affective lights. 

\ifCLASSOPTIONcaptionsoff
  \newpage
\fi

\bibliographystyle{IEEEtran}
\bibliography{main}

\begin{thebibliography}{10}
\providecommand{\url}[1]{#1}
\csname url@samestyle\endcsname
\providecommand{\newblock}{\relax}
\providecommand{\bibinfo}[2]{#2}
\providecommand{\BIBentrySTDinterwordspacing}{\spaceskip=0pt\relax}
\providecommand{\BIBentryALTinterwordstretchfactor}{4}
\providecommand{\BIBentryALTinterwordspacing}{\spaceskip=\fontdimen2\font plus
\BIBentryALTinterwordstretchfactor\fontdimen3\font minus
  \fontdimen4\font\relax}
\providecommand{\BIBforeignlanguage}[2]{{%
\expandafter\ifx\csname l@#1\endcsname\relax
\typeout{** WARNING: IEEEtran.bst: No hyphenation pattern has been}%
\typeout{** loaded for the language `#1'. Using the pattern for}%
\typeout{** the default language instead.}%
\else
\language=\csname l@#1\endcsname
\fi
#2}}
\providecommand{\BIBdecl}{\relax}
\BIBdecl

\bibitem{feng2021review}
Y.~Feng and S.~Cui, ``A review of emergency response in disasters: Present and
  future perspectives,'' \emph{Natural Hazards}, vol. 105, no.~1, pp.
  1109--1138, 2021.

\bibitem{shiri2020online}
D.~Shiri, V.~Akbari, and F.~S. Salman, ``Online routing and scheduling of
  search-and-rescue teams,'' \emph{OR Spectrum}, pp. 1--30, 2020.

\bibitem{adams2007search}
A.~L. Adams, T.~A. Schmidt, C.~D. Newgard, C.~S. Federiuk, M.~Christie,
  S.~Scorvo, and M.~DeFreest, ``Search is a time-critical event: when search
  and rescue missions may become futile,'' \emph{Wilderness \& Environmental
  Medicine}, vol.~18, no.~2, pp. 95--101, 2007.

\bibitem{slensky2004deployment}
K.~A. Slensky, K.~J. Drobatz, A.~B. Downend, and C.~M. Otto, ``Deployment
  morbidity among search-and-rescue dogs used after the september 11, 2001,
  terrorist attacks,'' \emph{Journal of the American Veterinary Medical
  Association}, vol. 225, no.~6, pp. 868--873, 2004.

\bibitem{miller200213}
G.~Miller, ``13 snake robots for search and rescue,'' \emph{Neurotechnology for
  Biomimetic Robots, MIT Press, Cambridge, MA}, p. 271, 2002.

\bibitem{hutson2017searching}
M.~Hutson, ``Searching for survivors of the mexico earthquake—with snake
  robots,'' \emph{Science}, 2017.

\bibitem{ye2006design}
C.~Ye, S.~Ma, and B.~Li, ``Design and basic experiments of a shape-shifting
  mobile robot for urban search and rescue,'' in \emph{2006 IEEE/RSJ
  International Conference on Intelligent Robots and Systems}.\hskip 1em plus
  0.5em minus 0.4em\relax IEEE, 2006, pp. 3994--3999.

\bibitem{Bethel2011}
C.~Bethel and R.~Murphy, ``{Non-Facial and Non-Verbal Affective Expression for
  Appearance-Constrained Robots Used in Victim Management*},'' \emph{Paladyn,
  Journal of Behavioral Robotics}, vol.~1, no.~4, pp. 219--230, 2011.

\bibitem{murphy2004trial}
R.~R. Murphy, ``Trial by fire [rescue robots],'' \emph{IEEE Robotics \&
  Automation Magazine}, vol.~11, no.~3, pp. 50--61, 2004.

\bibitem{pratt2006requirements}
K.~Pratt, R.~Murphy, S.~Stover, and C.~Griffin, ``Requirements for
  semi-autonomous flight in miniature uavs for structural inspection,''
  \emph{AUVSI's Unmanned Systems North America. Orlando, Florida, Association
  for Unmanned Vehicle Systems International}, 2006.

\bibitem{michael2014collaborative}
N.~Michael, S.~Shen, K.~Mohta, V.~Kumar, K.~Nagatani, Y.~Okada, S.~Kiribayashi,
  K.~Otake, K.~Yoshida, K.~Ohno \emph{et~al.}, ``Collaborative mapping of an
  earthquake damaged building via ground and aerial robots,'' in \emph{Field
  and service robotics}.\hskip 1em plus 0.5em minus 0.4em\relax Springer, 2014,
  pp. 33--47.

\bibitem{murphy2014disaster}
R.~R. Murphy, \emph{Disaster robotics}.\hskip 1em plus 0.5em minus 0.4em\relax
  MIT press, 2014.

\bibitem{matos2016multiple}
A.~Matos, A.~Martins, A.~Dias, B.~Ferreira, J.~M. Almeida, H.~Ferreira,
  G.~Amaral, A.~Figueiredo, R.~Almeida, and F.~Silva, ``Multiple robot
  operations for maritime search and rescue in eurathlon 2015 competition,'' in
  \emph{OCEANS 2016-Shanghai}.\hskip 1em plus 0.5em minus 0.4em\relax IEEE,
  2016, pp. 1--7.

\bibitem{casper2003human}
J.~Casper and R.~R. Murphy, ``Human-robot interactions during the
  robot-assisted urban search and rescue response at the world trade center,''
  \emph{IEEE Transactions on Systems, Man, and Cybernetics, Part B
  (Cybernetics)}, vol.~33, no.~3, pp. 367--385, 2003.

\bibitem{Gwaltney2003}
S.~Gwaltney-Brant, L.~Murphy, T.~Wismer, and J.~Albretsen, ``General
  toxicologic hazards and risks for search-and-rescue dogs responding to urban
  disasters,'' \emph{Journal of the American Veterinary Medical Association},
  vol. 222, pp. 292--5, 03 2003.

\bibitem{Linder2010}
T.~{Linder}, V.~{Tretyakov}, S.~{Blumenthal}, P.~{Molitor}, D.~{Holz},
  R.~{Murphy}, S.~{Tadokoro}, and H.~{Surmann}, ``Rescue robots at the collapse
  of the municipal archive of cologne city: A field report,'' in \emph{2010
  IEEE Safety Security and Rescue Robotics}, July 2010, pp. 1--6.

\bibitem{federal2013technical}
\BIBentryALTinterwordspacing
F.~E.~M. Agency and U.~Administration, \emph{Technical Rescue Program
  Development Manual}.\hskip 1em plus 0.5em minus 0.4em\relax CreateSpace
  Independent Publishing Platform, 2013. [Online]. Available:
  \url{https://books.google.ca/books?id=yVxLEcFoGoIC}
\BIBentrySTDinterwordspacing

\bibitem{delmerico2019current}
J.~Delmerico, S.~Mintchev, A.~Giusti, B.~Gromov, K.~Melo, T.~Horvat, C.~Cadena,
  M.~Hutter, A.~Ijspeert, D.~Floreano \emph{et~al.}, ``The current state and
  future outlook of rescue robotics,'' \emph{Journal of Field Robotics},
  vol.~36, no.~7, pp. 1171--1191, 2019.

\bibitem{kruijff2014experience}
G.-J.~M. Kruijff, M.~Jan{\'\i}{\v{c}}ek, S.~Keshavdas, B.~Larochelle,
  H.~Zender, N.~J. Smets, T.~Mioch, M.~A. Neerincx, J.~Diggelen, F.~Colas
  \emph{et~al.}, ``Experience in system design for human-robot teaming in urban
  search and rescue,'' in \emph{Field and Service Robotics}.\hskip 1em plus
  0.5em minus 0.4em\relax Springer, 2014, pp. 111--125.

\bibitem{jones2020}
B.~Jones, A.~Tang, and C.~Neustaedter, ``Remote communication in wilderness
  search and rescue: Implications for the design of emergency
  distributed-collaboration tools for network-sparse environments,''
  \emph{Proc. ACM Hum.-Comput. Interact.}, vol.~4, no. GROUP, Jan. 2020.

\bibitem{SARExample}
``The international maritime rescue federation mass rescue operations project:
  Communications – priorities, systems, structures,''
  \url{https://www.international-maritime-rescue.org/chapter-25-communications-priorities-systems-structures},
  accessed: 2022-05-24.

\bibitem{Liu2006}
J.~Liu, Y.~Wang, B.~Li, and S.~Ma, ``{Current research, key performances and
  future development of search and rescue robot},'' \emph{Jixie Gongcheng
  Xuebao/Chinese Journal of Mechanical Engineering}, vol.~42, no.~12, pp.
  1--12, 2006.

\bibitem{ekman1999basic}
P.~Ekman, ``Basic emotions,'' \emph{Handbook of cognition and emotion},
  vol.~98, no. 45-60, p.~16, 1999.

\bibitem{james1922emotions}
W.~James, \emph{What is an Emotion?}\hskip 1em plus 0.5em minus 0.4em\relax
  Simon and Schuster, 2013.

\bibitem{cannon1927james}
W.~B. Cannon, ``The james-lange theory of emotions: A critical examination and
  an alternative theory,'' \emph{The American journal of psychology}, vol.~39,
  no. 1/4, pp. 106--124, 1927.

\bibitem{Saraiva2019}
M.~Saraiva, H.~Ayano{\u{g}}lu, and B.~{\"O}zcan, ``Emotional design and
  human-robot interaction,'' in \emph{Emotional Design in Human-Robot
  Interaction}.\hskip 1em plus 0.5em minus 0.4em\relax Springer, 2019, pp.
  119--141.

\bibitem{darwin1872expression}
C.~Darwin, ``The expression of emotion in man and animals. london, england:
  Murray,'' 1872.

\bibitem{van2018}
B.~Van~Acker, D.~Parmentier, P.~Vlerick, and J.~Saldien, ``Understanding mental
  workload: from a clarifying concept analysis toward an implementable
  framework,'' \emph{Cognition, Technology \& Work}, 04 2018.

\bibitem{Kolling2016}
A.~Kolling, P.~Walker, N.~Chakraborty, K.~Sycara, and M.~Lewis, ``{Human
  Interaction with Robot Swarms: A Survey},'' \emph{IEEE Transactions on
  Human-Machine Systems}, vol.~46, no.~1, pp. 9--26, 2016.

\bibitem{marx2013rosen}
J.~Marx, R.~Walls, and R.~Hockberger, \emph{Rosen's emergency medicine-concepts
  and clinical practice}.\hskip 1em plus 0.5em minus 0.4em\relax Elsevier
  Health Sciences, 2013.

\bibitem{sefidgar2016}
Y.~S. {Sefidgar}, K.~E. {MacLean}, S.~{Yohanan}, H.~F.~M. {Van der Loos}, E.~A.
  {Croft}, and E.~J. {Garland}, ``Design and evaluation of a touch-centered
  calming interaction with a social robot,'' \emph{IEEE Transactions on
  Affective Computing}, vol.~7, no.~2, pp. 108--121, 2016.

\bibitem{bethel2010non}
C.~L. Bethel and R.~R. Murphy, ``Non-facial and non-verbal affective expression
  for appearance-constrained robots used in victim management*,''
  \emph{Paladyn, Journal of Behavioral Robotics}, vol.~1, no.~4, pp. 219--230,
  2010.

\bibitem{karaca2018potential}
Y.~Karaca, M.~Cicek, O.~Tatli, A.~Sahin, S.~Pasli, M.~F. Beser, and S.~Turedi,
  ``The potential use of unmanned aircraft systems (drones) in mountain search
  and rescue operations,'' \emph{The American journal of emergency medicine},
  vol.~36, no.~4, pp. 583--588, 2018.

\bibitem{jackovics2016standard}
P.~Jackovics, ``Standard of operation for cave rescue in hungary,''
  \emph{International Fire Fighter}, vol. 2016, no.~9, pp. 84--86, 2016.

\bibitem{baker2004improved}
M.~Baker, R.~Casey, B.~Keyes, and H.~A. Yanco, ``Improved interfaces for
  human-robot interaction in urban search and rescue,'' in \emph{2004 IEEE
  International Conference on Systems, Man and Cybernetics (IEEE Cat. No.
  04CH37583)}, vol.~3.\hskip 1em plus 0.5em minus 0.4em\relax IEEE, 2004, pp.
  2960--2965.

\bibitem{goodrich2008supporting}
M.~A. Goodrich, B.~S. Morse, D.~Gerhardt, J.~L. Cooper, M.~Quigley, J.~A.
  Adams, and C.~Humphrey, ``Supporting wilderness search and rescue using a
  camera-equipped mini uav,'' \emph{Journal of Field Robotics}, vol.~25, no.
  1-2, pp. 89--110, 2008.

\bibitem{1337826}
R.~R. {Murphy}, ``Trial by fire [rescue robots],'' \emph{IEEE Robotics
  Automation Magazine}, vol.~11, no.~3, pp. 50--61, 2004.

\bibitem{liu2013robotic}
Y.~Liu and G.~Nejat, ``Robotic urban search and rescue: A survey from the
  control perspective,'' \emph{Journal of Intelligent \& Robotic Systems},
  vol.~72, no.~2, pp. 147--165, 2013.

\bibitem{matsuno2014utilization}
F.~Matsuno, N.~Sato, K.~Kon, H.~Igarashi, T.~Kimura, and R.~Murphy,
  ``Utilization of robot systems in disaster sites of the great eastern japan
  earthquake,'' in \emph{Field and service robotics}.\hskip 1em plus 0.5em
  minus 0.4em\relax Springer, 2014, pp. 1--17.

\bibitem{mourikis2007autonomous}
A.~I. Mourikis, N.~Trawny, S.~I. Roumeliotis, D.~M. Helmick, and L.~Matthies,
  ``Autonomous stair climbing for tracked vehicles,'' \emph{The International
  Journal of Robotics Research}, vol.~26, no.~7, pp. 737--758, 2007.

\bibitem{finzi2005human}
A.~Finzi and A.~Orlandini, ``Human-robot interaction through mixed-initiative
  planning for rescue and search rovers,'' in \emph{Congress of the Italian
  Association for Artificial Intelligence}.\hskip 1em plus 0.5em minus
  0.4em\relax Springer, 2005, pp. 483--494.

\bibitem{wegner2006agent}
R.~Wegner and J.~Anderson, ``Agent-based support for balancing teleoperation
  and autonomy in urban search and rescue,'' \emph{International Journal of
  Robotics and Automation}, vol.~21, no.~2, pp. 120--128, 2006.

\bibitem{doroodgar2010hierarchical}
B.~Doroodgar and G.~Nejat, ``A hierarchical reinforcement learning based
  control architecture for semi-autonomous rescue robots in cluttered
  environments,'' in \emph{2010 IEEE International Conference on Automation
  Science and Engineering}.\hskip 1em plus 0.5em minus 0.4em\relax IEEE, 2010,
  pp. 948--953.

\bibitem{niroui2019deep}
F.~Niroui, K.~Zhang, Z.~Kashino, and G.~Nejat, ``Deep reinforcement learning
  robot for search and rescue applications: Exploration in unknown cluttered
  environments,'' \emph{IEEE Robotics and Automation Letters}, vol.~4, no.~2,
  pp. 610--617, 2019.

\bibitem{lygouras2019unsupervised}
E.~Lygouras, N.~Santavas, A.~Taitzoglou, K.~Tarchanidis, A.~Mitropoulos, and
  A.~Gasteratos, ``Unsupervised human detection with an embedded vision system
  on a fully autonomous uav for search and rescue operations,'' \emph{Sensors},
  vol.~19, no.~16, p. 3542, 2019.

\bibitem{anjomshoae2019}
S.~Anjomshoae, A.~Najjar, D.~Calvaresi, and K.~Fr\"{a}mling, ``Explainable
  agents and robots: Results from a systematic literature review,'' in
  \emph{Proceedings of the 18th International Conference on Autonomous Agents
  and MultiAgent Systems}, ser. AAMAS ’19.\hskip 1em plus 0.5em minus
  0.4em\relax Richland, SC: International Foundation for Autonomous Agents and
  Multiagent Systems, 2019, p. 1078–1088.

\bibitem{hellstrom2018}
T.~Hellström and S.~Bensch, ``Understandable robots-- what, why and how?''
  \emph{Paladyn, Journal of Behavioral Robotics}, vol.~9, pp. 110--123, 07
  2018.

\bibitem{rossi2017timing}
A.~Rossi, K.~Dautenhahn, K.~L. Koay, and M.~L. Walters, ``How the timing and
  magnitude of robot errors influence peoples’ trust of robots in an
  emergency scenario,'' in \emph{International Conference on Social
  Robotics}.\hskip 1em plus 0.5em minus 0.4em\relax Springer, 2017, pp. 42--52.

\bibitem{mirnig2017err}
N.~Mirnig, G.~Stollnberger, M.~Miksch, S.~Stadler, M.~Giuliani, and
  M.~Tscheligi, ``To err is robot: How humans assess and act toward an
  erroneous social robot,'' \emph{Frontiers in Robotics and AI}, vol.~4, p.~21,
  2017.

\bibitem{kleiner2007mapping}
A.~Kleiner, C.~Dornhege, and S.~Dali, ``Mapping disaster areas jointly:
  Rfid-coordinated slam by hurnans and robots,'' in \emph{2007 IEEE
  International Workshop on Safety, Security and Rescue Robotics}.\hskip 1em
  plus 0.5em minus 0.4em\relax IEEE, 2007, pp. 1--6.

\bibitem{chaffey2019developing}
P.~Chaffey, R.~Artstein, K.~Georgila, K.~A. Pollard, and S.~N. Gilani,
  ``Developing a virtual reality wildfire simulation to analyze human
  communication and interaction with a robotic swarm during emergencies,'' in
  \emph{Workshop on Human Language Technologies in Crisis and Emergency
  Management}, 2019.

\bibitem{sato2004cooperative}
N.~Sato, F.~Matsuno, T.~Yamasaki, T.~Kamegawa, N.~Shiroma, and H.~Igarashi,
  ``Cooperative task execution by a multiple robot team and its operators in
  search and rescue operations,'' in \emph{2004 IEEE/RSJ International
  Conference on Intelligent Robots and Systems (IROS)(IEEE Cat. No.
  04CH37566)}, vol.~2.\hskip 1em plus 0.5em minus 0.4em\relax IEEE, 2004, pp.
  1083--1088.

\bibitem{lewis2010teams}
M.~Lewis, H.~Wang, S.-Y. Chien, P.~Scerri, P.~Velagapudi, K.~Sycara, and
  B.~Kane, ``Teams organization and performance in multi-human/multi-robot
  teams,'' in \emph{2010 IEEE International Conference on Systems, Man and
  Cybernetics}.\hskip 1em plus 0.5em minus 0.4em\relax IEEE, 2010, pp.
  1617--1623.

\bibitem{hada2011}
Y.~Hada and O.~Takizawa, ``2-4 development of communication technology for
  search and rescue robots,'' \emph{Journal of the National Institute of
  Information and Communications Technology}, vol.~58, pp. 131--151, 03 2011.

\bibitem{mayer2019}
S.~Mayer, L.~Lischke, and P.~W. Wo{\'z}niak, ``{Drones for Search and
  Rescue},'' in \emph{{1st International Workshop on Human-Drone
  Interaction}}.\hskip 1em plus 0.5em minus 0.4em\relax Glasgow, United
  Kingdom: {Ecole Nationale de l'Aviation Civile [ENAC]}, May 2019.

\bibitem{fincannon2004evidence}
T.~Fincannon, L.~E. Barnes, R.~R. Murphy, and D.~L. Riddle, ``Evidence of the
  need for social intelligence in rescue robots,'' in \emph{2004 IEEE/RSJ
  International Conference on Intelligent Robots and Systems (IROS)(IEEE Cat.
  No. 04CH37566)}, vol.~2.\hskip 1em plus 0.5em minus 0.4em\relax IEEE, 2004,
  pp. 1089--1095.

\bibitem{murphy2004robot}
R.~R. Murphy, D.~Riddle, and E.~Rasmussen, ``Robot-assisted medical reachback:
  a survey of how medical personnel expect to interact with rescue robots,'' in
  \emph{RO-MAN 2004. 13th IEEE International Workshop on Robot and Human
  Interactive Communication (IEEE Catalog No. 04TH8759)}.\hskip 1em plus 0.5em
  minus 0.4em\relax IEEE, 2004, pp. 301--306.

\bibitem{breazeal2003emotion}
C.~Breazeal, ``Emotion and sociable humanoid robots,'' \emph{International
  journal of human-computer studies}, vol.~59, no. 1-2, pp. 119--155, 2003.

\bibitem{churamani2020icub}
N.~Churamani, F.~Cruz, S.~Griffiths, and P.~Barros, ``icub: learning emotion
  expressions using human reward,'' \emph{arXiv preprint arXiv:2003.13483},
  2020.

\bibitem{saldien2010expressing}
J.~Saldien, K.~Goris, B.~Vanderborght, J.~Vanderfaeillie, and D.~Lefeber,
  ``Expressing emotions with the social robot probo,'' \emph{International
  Journal of Social Robotics}, vol.~2, no.~4, pp. 377--389, 2010.

\bibitem{sosnowski2006design}
S.~Sosnowski, A.~Bittermann, K.~Kuhnlenz, and M.~Buss, ``Design and evaluation
  of emotion-display eddie,'' in \emph{2006 IEEE/RSJ International Conference
  on Intelligent Robots and Systems}.\hskip 1em plus 0.5em minus 0.4em\relax
  IEEE, 2006, pp. 3113--3118.

\bibitem{ghafurianMiro}
M.~Ghafurian, G.~Lakatos, Z.~Tao, and K.~Dautenhahn, ``Design and evaluation of
  affective expressions of a zoomorphic robot,'' in \emph{International
  Conference on Social Robotics}.\hskip 1em plus 0.5em minus 0.4em\relax
  Springer, 2020, pp. 1--12.

\bibitem{ghafurian2021recognition}
M.~Ghafurian, S.~A. Akgun, M.~Crowley, and K.~Dautenhahn, ``Recognition of a
  robot's affective expressions under conditions with limited visibility,'' in
  \emph{Human-Computer Interaction -- INTERACT 2021}, C.~Ardito, R.~Lanzilotti,
  A.~Malizia, H.~Petrie, A.~Piccinno, G.~Desolda, and K.~Inkpen, Eds.\hskip 1em
  plus 0.5em minus 0.4em\relax Cham: Springer International Publishing, 2021,
  pp. 448--469.

\bibitem{korcsok2018biologically}
B.~Korcsok, V.~Konok, G.~Persa, T.~Farag{\'o}, M.~Niitsuma, {\'A}.~Mikl{\'o}si,
  P.~Korondi, P.~Baranyi, and M.~G{\'a}csi, ``Biologically inspired emotional
  expressions for artificial agents,'' \emph{Frontiers in psychology}, vol.~9,
  p. 1191, 2018.

\bibitem{sharma2013communicating}
M.~Sharma, D.~Hildebrandt, G.~Newman, J.~E. Young, and R.~Eskicioglu,
  ``Communicating affect via flight path exploring use of the laban effort
  system for designing affective locomotion paths,'' in \emph{2013 8th ACM/IEEE
  International Conference on Human-Robot Interaction (HRI)}.\hskip 1em plus
  0.5em minus 0.4em\relax IEEE, 2013, pp. 293--300.

\bibitem{rea2012roomba}
D.~J. Rea, J.~E. Young, and P.~Irani, ``The roomba mood ring: an
  ambient-display robot,'' in \emph{Proceedings of the seventh annual ACM/IEEE
  international conference on Human-Robot Interaction}, 2012, pp. 217--218.

\bibitem{kayukawa2017influence}
Y.~Kayukawa, Y.~Takahashi, T.~Tsujimoto, K.~Terada, and H.~Inoue, ``Influence
  of emotional expression of real humanoid robot to human decision-making,'' in
  \emph{2017 IEEE International Conference on Fuzzy Systems (FUZZ-IEEE)}.\hskip
  1em plus 0.5em minus 0.4em\relax IEEE, 2017, pp. 1--6.

\bibitem{collins2015saying}
E.~C. Collins, T.~J. Prescott, and B.~Mitchinson, ``Saying it with light: A
  pilot study of affective communication using the miro robot,'' in
  \emph{Conference on Biomimetic and Biohybrid Systems}.\hskip 1em plus 0.5em
  minus 0.4em\relax Springer, 2015, pp. 243--255.

\bibitem{song2017expressing}
S.~Song and S.~Yamada, ``Expressing emotions through color, sound, and
  vibration with an appearance-constrained social robot,'' in \emph{2017 12th
  ACM/IEEE International Conference on Human-Robot Interaction (HRI}.\hskip 1em
  plus 0.5em minus 0.4em\relax IEEE, 2017, pp. 2--11.

\bibitem{andreasson2018affective}
R.~Andreasson, B.~Alenljung, E.~Billing, and R.~Lowe, ``Affective touch in
  human--robot interaction: conveying emotion to the nao robot,''
  \emph{International Journal of Social Robotics}, vol.~10, no.~4, pp.
  473--491, 2018.

\bibitem{Thelwall2012SentimentSD}
M.~Thelwall, K.~Buckley, and G.~Paltoglou, ``Sentiment strength detection for
  the social web,'' \emph{Journal of the American Society for Information
  Science and Technology}, vol.~63, no.~1, pp. 163--173, 2012.

\bibitem{turney2002thumbs}
P.~D. Turney, ``Thumbs up or thumbs down?: semantic orientation applied to
  unsupervised classification of reviews,'' in \emph{Proceedings of the 40th
  annual meeting on association for computational linguistics}.\hskip 1em plus
  0.5em minus 0.4em\relax Association for Computational Linguistics, 2002, pp.
  417--424.

\bibitem{ghazi2015detecting}
D.~Ghazi, D.~Inkpen, and S.~Szpakowicz, ``Detecting emotion stimuli in
  emotion-bearing sentences,'' in \emph{International Conference on Intelligent
  Text Processing and Computational Linguistics}.\hskip 1em plus 0.5em minus
  0.4em\relax Springer, 2015, pp. 152--165.

\bibitem{russell2015real}
E.~W. Russell, ``Real-time topic and sentiment analysis in human-robot
  conversation,'' Ph.D. dissertation, Marquette University, 2015.

\bibitem{mishra2019can}
N.~Mishra, M.~Ramanathan, R.~Satapathy, E.~Cambria, and N.~Magnenat-Thalmann,
  ``Can a humanoid robot be part of the organizational workforce? a user study
  leveraging sentiment analysis,'' in \emph{2019 28th IEEE International
  Conference on Robot and Human Interactive Communication (RO-MAN)}.\hskip 1em
  plus 0.5em minus 0.4em\relax IEEE, 2019, pp. 1--7.

\bibitem{nazir2020issues}
A.~Nazir, Y.~Rao, L.~Wu, and L.~Sun, ``Issues and challenges of aspect-based
  sentiment analysis: A comprehensive survey,'' \emph{IEEE Transactions on
  Affective Computing}, 2020.

\bibitem{mehrabian1995framework}
A.~Mehrabian, ``Framework for a comprehensive description and measurement of
  emotional states.'' \emph{Genetic, social, and general psychology
  monographs}, 1995.

\bibitem{heise2007expressive}
D.~R. Heise, \emph{Expressive order: Confirming sentiments in social
  actions}.\hskip 1em plus 0.5em minus 0.4em\relax Springer Science \& Business
  Media, 2007.

\bibitem{Akgun2020}
S.~A. Akgun, M.~Ghafurian, M.~Crowley, and K.~Dautenhahn, ``Using emotions to
  complement multi-modal human-robot interaction in urban search and rescue
  scenarios,'' in \emph{Proceedings of the 2020 International Conference on
  Multimodal Interaction}, ser. ICMI '20.\hskip 1em plus 0.5em minus
  0.4em\relax New York, NY, USA: Association for Computing Machinery, 2020, p.
  575–584.

\bibitem{benesty2009pearson}
J.~Benesty, J.~Chen, Y.~Huang, and I.~Cohen, ``Pearson correlation
  coefficient,'' in \emph{Noise reduction in speech processing}.\hskip 1em plus
  0.5em minus 0.4em\relax Springer, 2009, pp. 1--4.

\bibitem{smith2016mean}
L.~Smith-Lovin, D.~T. Robinson, B.~C. Cannon, J.~K. Clark, R.~Freeland, J.~H.
  Morgan, and K.~B. Rogers, ``Mean affective ratings of 929 identities, 814
  behaviors, and 660 modifiers by university of georgia and duke university
  undergraduates and by community members in durham, nc, in 2012-2014,''
  \emph{University of Georgia: Distributed at UGA Affect Control Theory
  Website: http://research. franklin. uga. edu/act}, 2016.

\bibitem{rafferty2013autonomous}
K.~Rafferty and E.~W. McGookin, ``An autonomous air-sea rescue system using
  particle swarm optimization,'' in \emph{2013 International Conference on
  Connected Vehicles and Expo (ICCVE)}.\hskip 1em plus 0.5em minus 0.4em\relax
  IEEE, 2013, pp. 459--464.

\bibitem{hung2007search}
E.~K. Hung and D.~A. Townes, ``Search and rescue in yosemite national park: a
  10-year review,'' \emph{Wilderness \& environmental medicine}, vol.~18,
  no.~2, pp. 111--116, 2007.

\bibitem{mu2020optimization}
L.~Mu and E.~Zhao, ``The optimization of maritime search and rescue simulation
  system based on cps,'' in \emph{Big Data Analytics for Cyber-Physical
  Systems}.\hskip 1em plus 0.5em minus 0.4em\relax Springer, 2020, pp.
  231--245.

\bibitem{allouche2010multi}
M.~K. Allouche and A.~Boukhtouta, ``Multi-agent coordination by temporal plan
  fusion: Application to combat search and rescue,'' \emph{Information Fusion},
  vol.~11, no.~3, pp. 220--232, 2010.

\bibitem{van2017wilderness}
C.~Van~Tilburg, C.~K. Grissom, K.~Zafren, S.~McIntosh, M.~I. Radwin, P.~Paal,
  P.~Haegeli, A.~R. Wheeler, D.~Weber, B.~Tremper \emph{et~al.}, ``Wilderness
  medical society practice guidelines for prevention and management of
  avalanche and nonavalanche snow burial accidents,'' \emph{Wilderness \&
  environmental medicine}, vol.~28, no.~1, pp. 23--42, 2017.

\bibitem{danielsson1980euclidean}
P.-E. Danielsson, ``Euclidean distance mapping,'' \emph{Computer Graphics and
  image processing}, vol.~14, no.~3, pp. 227--248, 1980.

\bibitem{castano2013definingZalgo}
C.~M. Casta{\~n}o~D{\'\i}az, ``Defining and characterizing the concept of
  internet meme,'' \emph{Ces Psicolog{\'\i}a}, vol.~6, no.~2, pp. 82--104,
  2013.

\bibitem{motorola7550}
R.~Networks, ``Gps enabled two way radios make difference with search teams.''

\bibitem{wilkins2010led}
A.~Wilkins, J.~Veitch, and B.~Lehman, ``Led lighting flicker and potential
  health concerns: Ieee standard par1789 update,'' in \emph{2010 IEEE Energy
  Conversion Congress and Exposition}.\hskip 1em plus 0.5em minus 0.4em\relax
  IEEE, 2010, pp. 171--178.

\bibitem{quigley2009ros}
M.~Quigley, K.~Conley, B.~Gerkey, J.~Faust, T.~Foote, J.~Leibs, R.~Wheeler, and
  A.~Ng, ``Ros: an open-source robot operating system,'' in \emph{ICRA Workshop
  on Open Source Software}, vol.~3, 01 2009.

\bibitem{bates2005fitting}
D.~Bates, ``Fitting linear mixed models in r,'' \emph{R news}, vol.~5, no.~1,
  pp. 27--30, 2005.

\bibitem{arnold2010uninformative}
T.~W. Arnold, ``Uninformative parameters and model selection using akaike's
  information criterion,'' \emph{The Journal of Wildlife Management}, vol.~74,
  no.~6, pp. 1175--1178, 2010.

\bibitem{cohen2014context}
R.~Cohen and A.~W. Siegel, ``A context for context: Toward an analysis of
  context and development,'' in \emph{Context and development}.\hskip 1em plus
  0.5em minus 0.4em\relax Psychology Press, 2014, pp. 13--34.

\bibitem{ma2018socio}
X.~Ma, M.~Tamir, and Y.~Miyamoto, ``A socio-cultural instrumental approach to
  emotion regulation: Culture and the regulation of positive emotions.''
  \emph{Emotion}, vol.~18, no.~1, p. 138, 2018.

\bibitem{seppanen2013developing}
H.~Sepp{\"a}nen, J.~M{\"a}kel{\"a}, P.~Luokkala, and K.~Virrantaus,
  ``Developing shared situational awareness for emergency management,''
  \emph{Safety science}, vol.~55, pp. 1--9, 2013.

\bibitem{goodman2013data}
J.~K. Goodman, C.~E. Cryder, and A.~Cheema, ``Data collection in a flat world:
  The strengths and weaknesses of mechanical turk samples,'' \emph{Journal of
  Behavioral Decision Making}, vol.~26, no.~3, pp. 213--224, 2013.

\bibitem{10.1145/3405450}
D.~Feil-Seifer, K.~S. Haring, S.~Rossi, A.~R. Wagner, and T.~Williams, ``{Where
  to Next? The Impact of COVID-19 on Human-Robot Interaction Research},''
  \emph{ACM Transactions on Human-Robot Interaction}, vol.~10, no.~1, 6 2020.

\end{thebibliography}

%
\vskip -0.45in
\begin{IEEEbiography}[{\includegraphics[width=1in,height=1.25in,clip,keepaspectratio]{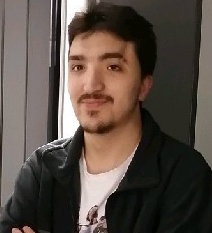}}]{Sami Alperen Akgun}
is a MAsc student in the Systems Design Engineering Department at the University of Waterloo and a member of Social and Intelligent Robotics Research Laboratory (SIRRL). His background is in control theory and automation and his research interests are robotics, eXplainable AI and HRI. His research at SIRRL focuses on human-robot interaction in search and rescue context.
\end{IEEEbiography}

\vskip -0.4in

\begin{IEEEbiography}[{\includegraphics[width=1in,height=1.25in,clip,keepaspectratio]{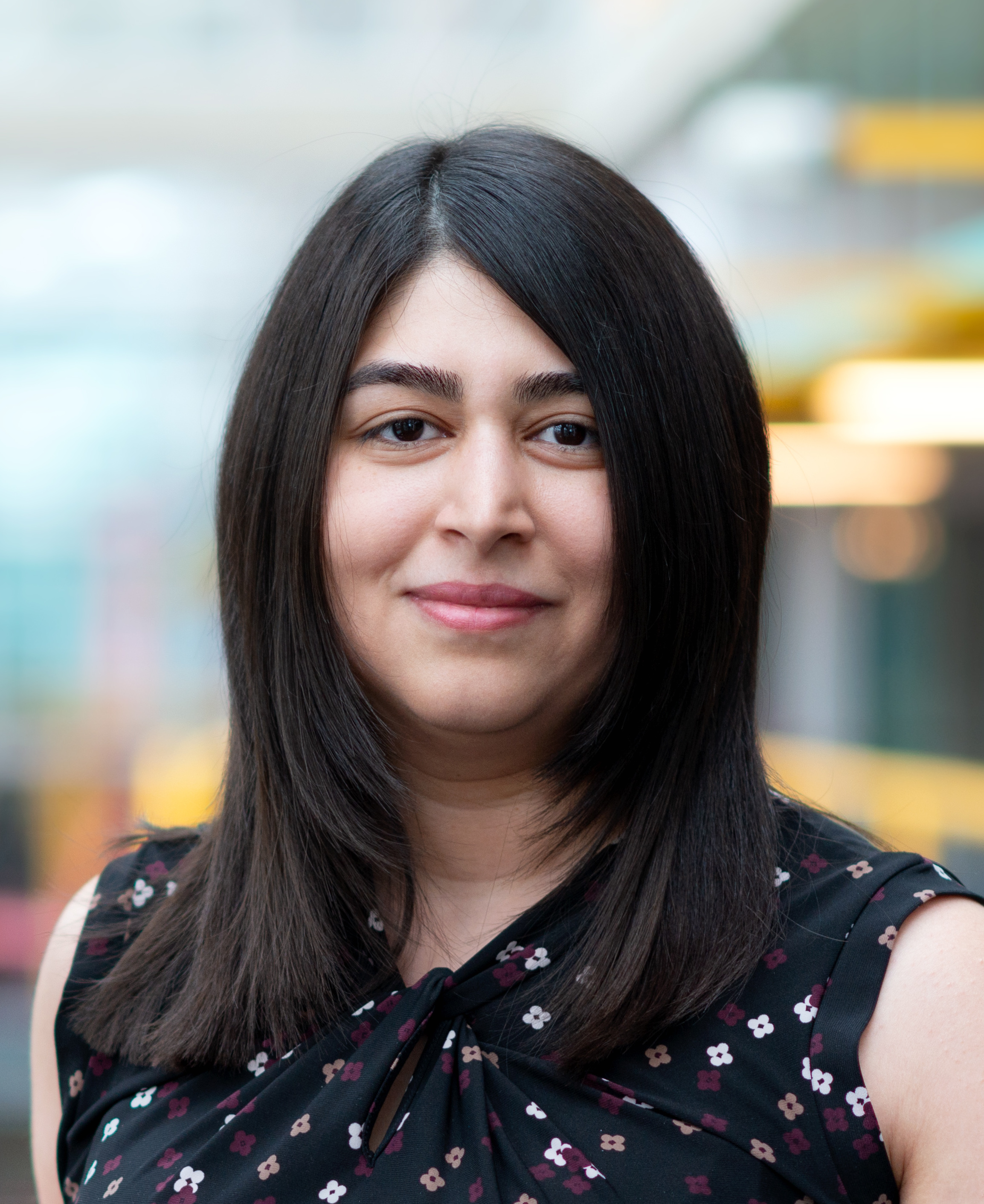}}]{Moojan Ghafurian}
is a Research Assistant Professor in the Department of Electrical and Computer Engineering at the University of Waterloo. She got her PhD from the Pennsylvania State University and was the Inaugural Wes Graham postdoctoral fellow from 2018-2020 at David R. Cheriton School of Computer Science at the University of Waterloo. Her research areas are human-computer/robot interaction, social robotics, affective computing, and cognitive science. Her research explores computational models of how humans interact with systems to inform user-centered design of emotionally intelligent agents in multiple domains.  
\end{IEEEbiography}
\vskip -0.4in

\begin{IEEEbiography}[{\includegraphics[width=1in,height=1.25in,clip,keepaspectratio]{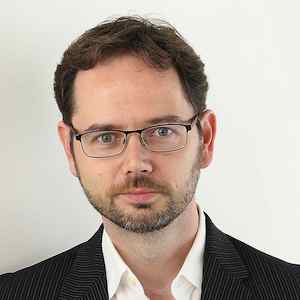}}]{Mark Crowley}
Mark Crowley is an Assistant Professor in the Department of Electrical and Computer Engineering at the University of Waterloo. He carries out research to find dependable and transparent ways to augment human decision making in complex domains, especially in the presence of spatial structure, streaming data, and uncertainty.  His research is applied to areas such as autonomous driving, human-robot collaboration, Sustainable Forest Management and Medical Imaging. His work falls into the fields of Reinforcement Learning, Multi-agent systems, and Manifold Learning.
\end{IEEEbiography}

\vskip -0.4in
\begin{IEEEbiography}[{\includegraphics[width=1in,height=1.25in,clip,keepaspectratio]{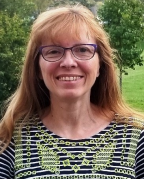}}]{Kerstin Dautenhahn}
is Canada 150 Research Chair in Intelligent Robotics in the Faculty of Engineering at University of Waterloo in Ontario, Canada where she directs the Social and Intelligent Robotics Research lab. She became IEEE Fellow in 2019 for her contributions to Social Robotics and Human-Robot Interaction. From 2000-2018 she coordinated the Adaptive Systems Research Group at University of Hertfordshire, UK. Her main research areas are human-robot interaction, social robotics, cognitive and developmental robotics, and assistive technology.
\end{IEEEbiography}





\vfill


\end{document}